\documentclass{article} 
\usepackage{iclr2027_conference,times}

\usepackage{amsmath,amsfonts,bm}

\def\eqref#1{equation~\ref{#1}}

\def\1{\bm{1}}

\DeclareMathAlphabet{\mathsfit}{\encodingdefault}{\sfdefault}{m}{sl}
\SetMathAlphabet{\mathsfit}{bold}{\encodingdefault}{\sfdefault}{bx}{n}

\usepackage{hyperref}
\usepackage{url}
\usepackage{booktabs}       
\usepackage{amsfonts}       
\usepackage{nicefrac}       
\usepackage{microtype}      
\usepackage{xcolor}         
\usepackage{amsmath}
\usepackage{amsthm}

\usepackage{algorithm}
\usepackage{algpseudocode}
\usepackage{graphicx}
\usepackage{subcaption}

\title{Inferring Multi-Timescale Neural Dynamics with Switching Linear Dynamical Systems}

\author{Lulu Gong,  Yongxu Zhang,   Shreya Saxena \\
Center for Neurocomputation and Machine Intelligence, Wu Tsai Institute\\
Department of Biomedical Engineering\\
Yale University\\
New Haven, CT 06510, USA \\
\texttt{\{lulu.gong\}@yale.edu} 
}

\iclrfinalcopy 
\begin{document}

\maketitle

\begin{abstract}
Neural activity often exhibits multiple timescales that can vary with behavioral states and task conditions. Identifying these timescales from neural recordings is important for better understanding neural computation and function. However, traditional approaches based on autocorrelation fitting are difficult to scale to high-dimensional population recordings and can become unreliable when neural dynamics change with behavior. State-space models have been a powerful framework for modeling high-dimensional neural population activity through latent dynamical systems, but standard formulations and inference methods do not explicitly account for multiple timescales and therefore do not guarantee accurate recovery of the underlying temporal structure. Motivated by these questions, we introduce the Multi-Timescale Switching Linear Dynamical System (MTS-SLDS), a framework for identifying regime-specific latent timescales from continuous or spiking neural observations. MTS-SLDS combines a multi-lag moment initialization, which captures temporal structure across multiple observation lags, with \textit{regime-conditioned} Laplace-EM inference, which reduces mixing of dynamical statistics across uncertain regimes. Characteristic timescales can then be extracted directly from the eigenvalues of the learned latent transition matrices. In synthetic experiments with Gaussian and Poisson observations, MTS-SLDS  accurately recovers timescales and switching structure over a range of signal levels. 
We further apply this framework on neural spike recordings from two behavioral tasks. During a fixation task with neural activity recorded from V4, the inferred timescales agree with estimates from a dedicated Bayesian autocorrelation method. In Area 2 motor cortex recordings during a reaching task, we find that MTS-SLDS identifies behaviorally aligned dynamical regimes and provides more interpretable regime-specific timescale estimates.  
These results establish MTS-SLDS as a reliable framework for investigating how neural timescales behave within and across behavioral states, and for studying their role in neural population dynamics.

\end{abstract}

\section{Introduction}
\label{sec:introduction}
Neuroscience experiments have revealed that 
activity from the brain evolves over multiple timescales, which reflects how neural circuits integrate information and maintain temporally extended representations. Neural recordings exhibit a hierarchy of temporal correlations across cortical areas, with generally shorter timescales in sensory regions and longer timescales in association cortex \citep{murray2014hierarchy,soltani2021timescales,spitmaan2020multiple,shi2025brain}. Multiple timescales also coexist within individual populations and change with attention and working-memory demands \citep{zeraati2023intrinsic,gao2020neuronal,zeraati2026neural}. These observations connect temporal organization to neural computation: fast dynamics support sensitivity to changing inputs, whereas slower dynamics support integration and persistence \citep{chaudhuri2015large}. Recovering this temporal structure can therefore reveal computational differences that average firing rates or trajectory visualizations alone may obscure.

Current approaches to recovering neural timescales are largely based on statistical summaries: the autocorrelation function of a recorded signal, or of a population average, is fitted along with one or more exponential decays, and the fitted decay constants are reported as timescales \citep{murray2014hierarchy, zeraati2022flexible}. These estimates depend on observation noise, on which projection of the population is measured, and on the assumed form of the decay; importantly, they carry no description of the dynamics that produced the correlations. Latent variable models of neural activity offer a different route \citep{paninski2018neural}. They posit that the recorded population is driven by a lower-dimensional latent state, $x_t$, that evolves according to its own dynamics, and is observed through a measurement process, such as a Gaussian readout of continuous signals or Poisson spike counts. Because this is a generative model rather than a summary statistic, it provides a richer dynamical description of the population, supports model-based prediction and held-out validation, and separates the dynamics of interest from the noise through which they are observed. 

In a linear dynamical system, the latent state evolves across successive time bins of width $\Delta t$ as $x_{t} = A x_{t-1} + w_t$, and in a switching linear dynamical system (SLDS) a discrete regime $z_t$ selects among transition matrices $A_1,\dots,A_K$ so that the dynamics can change with behavioral state. The eigenvalues of each recovered $A_k$ then directly inform us of the timescales: a stable eigenvalue $\lambda$ decays by a factor $|\lambda|$ per bin, which corresponds to a relaxation time of $-\Delta t/\log|\lambda|$. However, neural recordings are usually limited in duration and trial count, and with limited data the recovered dynamics are not uniquely determined. Two
transition matrices with different eigenvalues can yield similar likelihoods,
because the likelihood is most sensitive to the predicted observations and only indirectly to the decay rates behind them. Different fitted models can therefore reconstruct similar latent trajectories while assigning different decay rates or different switching structure to the underlying dynamics. This identifiability problem makes it difficult to interpret the eigenvalues of a
fitted model as timescales, which matters when the scientific objective is to interpret the learned system rather than only predict its observations.

Here, we introduce the Multi-Timescale Switching Linear Dynamical System (MTS-SLDS), a learning framework for recovering regime-specific latent timescales from multivariate Gaussian or Poisson observations. 
MTS-SLDS addresses the problem with two components. Firstly, a multi-lag moment initialization estimates the latent dynamics from observation moments at several lags, using moment relations specific to the Gaussian or Poisson observation model. Secondly, regime-conditioned Laplace EM (RC-L-EM) refines the model while retaining regime-specific state moments for $A_k$ updates, so that uncertain regime assignments do not mix dynamical statistics across regimes. 

We evaluate MTS-SLDS against a standard SLDS in stationary and switching simulations with Gaussian and Poisson observations, across observation noise levels and mean spike counts per bin. Timescale recovery is measured directly against the ground-truth spectrum alongside latent-state reconstruction and regime identification, and the two can diverge: models with similar reconstruction accuracy can differ substantially in the timescales they recover. We then apply the framework to two different neural recordings. In V4 during sustained fixation, a single-regime model yields timescales consistent with autocorrelation-based estimates, with higher held-out likelihood and closer simulated autocorrelations than the SLDS baseline. In somatosensory area 2 during reaching, the inferred regimes align with movement onset and the regime-specific timescales agree with behavioral decay times. MTS-SLDS thus offers a principled route to characterizing how neural timescales differ across behavioral states, with the recovered eigenvalues carrying a direct biological interpretation.

\subsection{Related work}

\paragraph{Timescale estimation}
Exponential function fits to neural recording autocorrelation have been widely used to discover cortical temporal organization \citep{murray2014hierarchy,soltani2021timescales,spitmaan2020multiple,cavanagh2016autocorrelation}. 
Frequency-domain approaches provide complementary estimates from
aperiodic spectral knees while separating oscillatory peaks from the
aperiodic background \citep{donoghue2020parameterizing,gao2020neuronal}.  Adaptive approximate Bayesian computation (aABC) improves on direct fitting by comparing simulated and observed summary statistics, addressing finite-sample bias and quantifying uncertainty within candidate generative models \citep{zeraati2022flexible}. When applied to population-summed counts, both approaches characterize a scalar projection whose weighting of latent modes can obscure some temporal components. Autoregressive models can represent multiple decay rates through their poles \citep{lutkepohl2005new}, while multistep regression can address subsampling bias under appropriate model assumptions \citep{wilting2018inferring}. A stationary autoregressive fit alone, however, does not explicitly separate latent variability from observation noise or infer changes between dynamical regimes.

\paragraph{Spectral and switching state-space methods}
Subspace identification connects temporal observation structure to
state-space realizations \citep{overschee1996subspace}.
Spectral learning connects observation moments to latent dynamics, including moment conversion for Poisson observations \citep{buesing2012spectral}. More recent methods identify dynamics shared between generalized-linear
processes and explicitly evaluate dynamical eigenvalue recovery
\citep{oganesian2024spectral}. SLDSs and recurrent extensions represent complex neural activity through locally linear regimes \citep{linderman2017bayesian,zoltowski2020general}. Their learning procedures differ in how they approximate uncertainty: factorized variational Laplace EM uses shared continuous-state moments weighted by regime responsibilities, whereas switching filtering approaches retain regime-conditioned estimates. Related methods support heterogeneous neural observations and Poisson cubature filtering, including sPCF \citep{song2022modeling,song2023unsupervised}. These developments establish important foundations, but neither model flexibility nor a particular posterior approximation alone ensures spectral recovery. MTS-SLDS builds on them by combining regime-local multi-lag initialization with regime-conditioned learning and explicitly testing whether the inferred dynamics recover the underlying timescales.

\section{Multi-Timescale Modeling and Inference}
\label{sec:mts_slds_model_method}

\subsection{Multiple timescales in neural dynamics}
\label{sec:mts_slds_timescales}

Neural population activity exhibits coexisting fast and slow fluctuations
\citep{murray2014hierarchy,zeraati2023intrinsic}. Their characteristic
timescales are commonly estimated from the decay of the autocorrelation
function (ACF). For a stationary scalar signal $s_t$ with positive variance, the
 ACF is defined
$\rho(\ell)=\operatorname{Cov}(s_{t+\ell},s_t)/\operatorname{Var}(s_t)$ with $\ell=1,2,\ldots$ being the lag in time bins. A common approximation is
\begin{equation}
\rho(\ell)\approx\sum_{j=1}^{M}a_j
\exp\!\left(-\frac{\ell\Delta t}{\tau_j}\right),
\qquad \ell>0,
\label{eq:neural_acf_timescales}
\end{equation}
where $\Delta t$ is the sampling interval, $a_j$ is the amplitude of
component $j$, and $\tau_j>0$ is its characteristic decay time.

To connect observed correlation decay to latent dynamics, consider
the linear state-space model
\[
x_{t+1}=Ax_t+w_t,
\qquad
y_t=Cx_t+\epsilon_t,
\]
where $A$ governs latent evolution, $C$ maps latent states to
observations, and $\eta_t$ and $\epsilon_t$ are mutually independent,
zero-mean white noise processes. For stable, stationary dynamics,
the observation covariance at positive lags satisfies
\[
\operatorname{Cov}(y_{t+\ell},y_t)
=
CA^\ell\Sigma_xC^\top,
\qquad \ell=1,2,\ldots,
\]
where $\Sigma_x$ is the stationary latent covariance. Thus, latent
dynamics shape observed correlations through powers of $A$.
Each eigenmode with eigenvalue $\lambda_j$ has relaxation timescale
\[
\tau_j=-\frac{\Delta t}{\log|\lambda_j|},
\qquad 0<|\lambda_j|<1,
\]
which characterizes its exponential decay envelope. This connection
motivates estimating latent timescales from fitted transition spectra
rather than directly fitting empirical ACFs. Moreover, spatial pooling operation in ACF based methods can also suppress some modal contributions, so the ACF
of a summed population signal may reveal only part of the latent
timescale spectrum (see Appendix~\ref{app:pooling}). Retaining multichannel
observations preserves spatial information that this scalar
summary can discard.
Appendix~\ref{app:timescale_theory} provides the detailed derivation
and explains how mode observability, nonlinear observations, and
regime switching affect the correspondence with observed
correlation decay.

\subsection{Multi-timescale SLDS model}
\label{sec:mts_slds_model}
We model neural population activity using a switching linear dynamical
system (SLDS), with continuous latent state $x_t\in\mathbb R^H$
and discrete regime $z_t\in\{1,\ldots,K\}$.
The model
evolves as
\begin{align}
z_t\mid z_{t-1}=i
&\sim\operatorname{Cat}(P_{i,:}),\nonumber\\
x_t\mid x_{t-1},z_t=k,u_t
&\sim\mathcal N(A_kx_{t-1}+b_k,Q_k),
\label{eq:mts_slds_dynamics}
\end{align}
 where $P$ is the regime transition matrix, and each regime has
a latent transition matrix $A_k\in\mathbb R^{H\times H}$,
offset $b_k$, and process-noise covariance $Q_k$.

 For continuous
measurements $y_t\in\mathbb R^D$, we use Gaussian emission model
\begin{equation}
y_t\mid x_t,z_t=k\sim\mathcal N(C_kx_t+d_k,R_k).
\label{eq:mts_slds_gaussian_obs}
\end{equation}
For spike counts, we assume conditional independence across neurons
and use Poisson emission model
\begin{equation}
y_{nt}\mid x_t,z_t=k
\sim\operatorname{Poisson}\!\left(\exp(c_{n,k}^{\top}x_t+d_{n,k})\right),
\qquad n=1,\ldots,D,
\label{eq:mts_slds_poisson_obs}
\end{equation}
where $c_{n,k}^{\top}$ is row $n$ of $C_k$, and the Poisson
parameter is the expected spike count per bin.

We characterize each regime through the eigenvalues
$\lambda_{k,j}=\lambda_j(A_k)$, which may be real or occur
in complex-conjugate pairs. For each decaying mode,
$0<|\lambda_{k,j}|<1$, its relaxation timescale and
oscillation frequency are
\begin{equation}
\tau_{k,j}
=-\frac{\Delta t}{\log|\lambda_{k,j}|},
\qquad
f_{k,j}
=\frac{|\arg(\lambda_{k,j})|}{2\pi\Delta t},
\label{eq:mts_slds_timescale}
\end{equation}
where $\Delta t$ is the sampling interval and $\arg$ denotes
the principal phase angle. Positive real eigenvalues describe
nonoscillatory decay, whereas complex-conjugate pairs describe
damped oscillations with a shared decay timescale and frequency. This spectral description allows multiple relaxation timescales
to coexist within each regime. These timescales characterize
latent dynamics with the regime held fixed and are distinct
from the regime dwell times governed by $P$
(see Appendix~\ref{app:switching_timescales}).

\begin{figure}[htbp]
    \centering
    \includegraphics[width=0.7\linewidth]{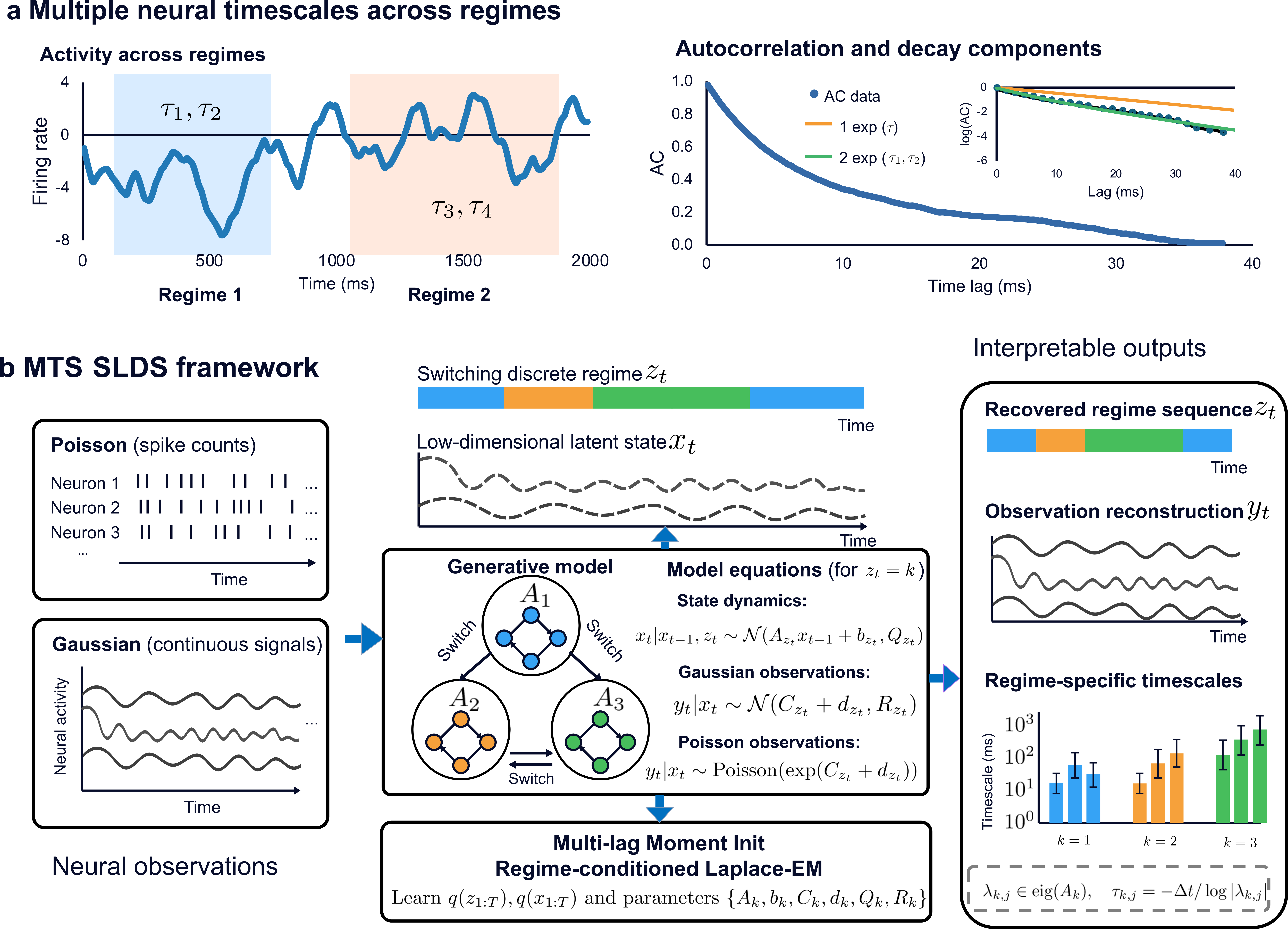}
    \caption{\textbf{Multiple neural timescales and the MTS-SLDS framework.} \textbf{A} Neural activity can switch between regimes with different characteristic timescales (here $\tau_1,\tau_2$ and $\tau_3,\tau_4$). The pooled autocorrelation of such a signal is not well described by a single exponential, and a multi-exponential fit summarizes the decay without attributing components to regimes. \textbf{B} MTS-SLDS models Poisson spike counts or Gaussian signals as observations of a latent state $x_t$ whose linear dynamics $A_{z_t}$ are selected by a discrete regime $z_t$. Multi-lag moment initialization and regime-conditioned Laplace EM (RC-L-EM) learn the regime sequence, latent}
    \label{fig:mts-slds}
\end{figure}

\subsection{Learning regime-specific timescales}
\label{sec:mts_slds_learning}

We propose a unified framework \emph{MTS-SLDS} to recover timescales in \eqref{eq:mts_slds_dynamics}, \ref{eq:mts_slds_gaussian_obs}, and \ref{eq:mts_slds_poisson_obs} through a two-stage procedure of
multi-lag initialization and regime-conditioned learning (Figure~\ref{fig:mts-slds} summarizes the whole framework).
MTS-SLDS first estimates initial dynamics and a provisional segmentation
from multi-lag observation moments. It then refines the model using
regime-conditioned Laplace-EM (RC-L-EM), which conditions dynamical
statistics on candidate regimes. After fitting, we extract timescales
from each $A_k$ using ~\eqref{eq:mts_slds_timescale}. Our ablation experiments show that both stages contribute to more
accurate regime identification and timescale estimation
(Appendix~\ref{app:ablation}).

\subsubsection{Multi-lag moment initialization}
\label{sec:initialization}
We initialize dynamics by relating observation moments across
multiple time lags.
Consider a stable, stationary Gaussian latent reference process with
regime $k$ held fixed. Let $\Sigma_k$ denote its
latent covariance and define
$\Gamma_{y,k}(\ell)=\operatorname{Cov}(y_{t+\ell},y_t)$
for positive integer lags $\ell$.
For Gaussian observations, we set $M_{\ell,k}=\Gamma_{y,k}(\ell)$.
For Poisson observations with the exponential link, we use the element-wise conversion
\begin{equation}
[M_{\ell,k}]_{nm}
=\log\!\left(
1+\frac{[\Gamma_{y,k}(\ell)]_{nm}}{\mu_{n,k}\mu_{m,k}}
\right),
\qquad \ell>0,
\label{eq:mts_slds_count_lag}
\end{equation}
where $\mu_{n,k}=\mathbb E[y_{nt}]$ under the reference process.
Both observation models then satisfy
$M_{\ell,k}=C_kA_k^\ell\Sigma_kC_k^\top$ for $\ell>0$
\citep{buesing2012spectral}
(see Appendix~\ref{app:timescale_theory} for  the detailed relations).

In practice, we construct a low-dimensional projection from
pooled within-trial positive-lag moments and obtain provisional
regime labels by applying $K$-means to standardized windowed
autocorrelation descriptors. We then estimate regime-specific
observation means and recompute moments using only lag intervals
assigned entirely to the same regime. Regularized regression
between successive projected moment matrices across selected
lags yields initial transition operators, which we map to latent
coordinates compatible with the emission parameters (see Appendix~\ref{app:moment_initialization} for details). These stationary identities motivate the initialization but need
not hold exactly for empirically segmented switching data.
We therefore use the resulting operators as starting estimates
for RC-L-EM.

\subsubsection{Regime-conditioned Laplace EM (RC-L-EM)}
\label{sec:rclem}
In SLDS, variational Laplace EM with a factorized
posterior $q_x(x_{1:T})q_z(z_{1:T})$ has been widely used \citep{linderman2017bayesian,zoltowski2020general}.
Its chain-structured $q_z$ provides soft regime assignments, and
$q_x$ is a single Gaussian over the latent trajectory, shared by all regimes.
The update for $A_k$ therefore uses the same state moments for every $k$, weighted only by the responsibility of regime $k$; under uncertain assignments, each $A_k$ is fitted to moments that can partly reflect the other regimes.
RC-L-EM instead maintains a separate Gaussian state approximation for each candidate regime and adjacent regime pair through switching filtering and smoothing \citep{song2022modeling,song2023unsupervised}.
Appendix~\ref{app:ssm_baseline} details the variational baseline. RC-L-EM has the property that the moments entering the $A_k$ update are computed assuming regime $k$ generated the transition.

During the E-step, the forward filter represents
\begin{equation}
q_f(x_t,z_t=k)
=\gamma_t^f(k)\mathcal N(x_t;m_t^{(k)},V_t^{(k)}),
\label{eq:rclem_filter}
\end{equation}
where $\gamma_t^f(k)\approx p(z_t=k\mid y_{1:t})$ and $(m_t^{(k)},V_t^{(k)})$ approximate the state given $z_t=k$, so each regime carries its own state estimate.
For each destination regime, we match the first two moments of
incoming state components, propagate them through $A_k$,
and incorporate $y_t$.
Gaussian observations use a Gaussian measurement update; Poisson
observations use a local Laplace approximation for each $k$.
The predictive evidence
$\mathcal L_t(k)\approx p(y_t\mid z_t=k,y_{1:t-1})$
combines with the Markov prediction to update the regime weights.
Appendix~\ref{app:rclem_filtering} gives these calculations.

Backward smoothing leads to regime probabilities $\gamma_t^s(k)$
and adjacent-regime probabilities $\xi_t(i,k)$ conditioned on all
observations. It also provides Gaussian pair approximations
$q_t^{ik}(x_{t-1},x_t)$ conditional on $z_{t-1}=i,z_t=k$ and
$y_{1:T}$. The two methods supply the cross-moment for the $A_k$ update differently: 
\begin{equation}
M_{t,k}^{\mathrm{RC}}
=\sum_i\xi_t(i,k)
\mathbb E_{q_t^{ik}}[x_tx_{t-1}^{\top}] \qquad\text{versus}\qquad  
M_{t,k}^{\mathrm{MF}}
=\gamma_t^s(k)\mathbb E_{q_x}[x_tx_{t-1}^{\top}].
\label{eq:rclem_conditional_cross_moment}
\end{equation}
In the factorized form only the scalar weight depends on $k$; in RC-L-EM the expectation itself is conditioned on the regime pair, so the state mean and covariance, not only the weight, change with the candidate regime. Both methods use the same expected-regression formula and differ only in the moments supplied to it (Appendix~\ref{app:rclem_posterior}).

In the M-step, expected regression updates the linear dynamics,
expected transition counts update $P$, and regime-conditioned state
moments determine the emission updates.
Gaussian exponential moments give an analytic expected Poisson
log-likelihood. We optimize this objective numerically.
Appendices~\ref{app:rclem_mstep} and~\ref{app:rclem_observation_updates}
give the updates for each trainable parameter block.
We alternate the E- and M-steps and run final inference at the fitted
parameters. Moment matching and local Laplace updates make this an
approximate EM procedure.

\begin{figure}[h!]
    \centering
    \includegraphics[width=0.65\linewidth]{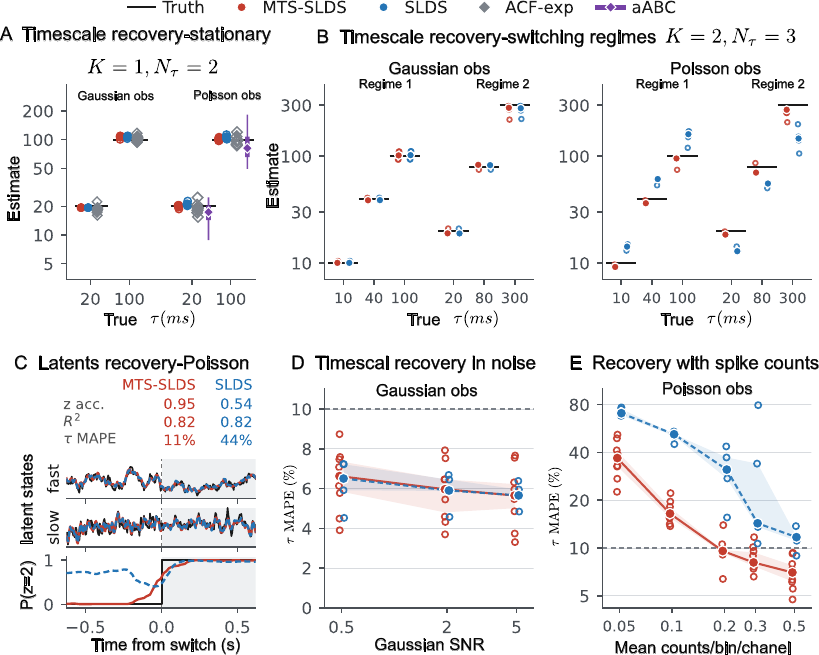}
    \caption{\textbf{Timescale recovery in stationary and switching systems with Gaussian and Poisson observations.} \textbf{A} Estimated timescales under stationary
($K=1$, $N_\tau=2$), where hollow markers represent different runs, and filled markers show medians. aABC shows the posterior median and $50\%$/$95\%$ credible
intervals. \textbf{B} Estimated timescales under switching
($K=2$, $N_\tau=3$ per regime) dynamics, where circles denote latent modes of timescales, and dots dentoe medians.
\textbf{C} Example held-out latent continuous states and regime inference in switching dynamics.
\textbf{D, E} Timescale MAPE versus Gaussian SNR
and mean Poisson counts in switching dynamics, respectively, where
shading denotes interquartile ranges and dashed lines denotes the reference $10\%$ error level.}
\label{fig:synthetic}
\end{figure}

\section{Synthetic data}
\label{sec:synthetic_results}
We evaluate timescale identification under known ground truth,
progressing from stationary dynamics to switching dynamics with multiple regimes. For each experiment, 
we generate 128-channel Gaussian or Poisson observations from
10-dimensional latent (switching) linear dynamical systems, using 5-ms bins and
30 training and 20 held-out trials of 4~s each.
We prescribe timescales through the transition eigenvalues,
$|\lambda_{k,j}|=\exp(-\Delta t/\tau_{k,j})$.
The stationary experiment ($K=1$) contains timescales of
$20$ and $100$~ms, each represented by five latent modes. In the switching experiment ($K=2$), we alternate timescales between spectra
$\{10,40,100\}$~ms and $\{20,80,300\}$~ms, with $3/3/4$ modes per timescale in each regime.
A symmetric Markov chain governs the regime switching, with
self-transition probability $0.9975$, corresponding to
an expected dwell time of $2$~s.

We compare MTS-SLDS with the standard SLDS (implementation in \citep{ssmsoftware}), using each method's default initialization and inference
procedure. In stationary comparisons, we additionally include
two-exponential autocorrelation fitting (ACF-exp) for both
observation models and adaptive approximate Bayesian computation
(aABC) for Poisson observations specifically.
These two methods use population-summed
spike counts in the Poisson comparison. After fitting, 
we extracted inferred timescales as
$\widehat{\tau}_{k,j}=-\Delta t/\log|\widehat{\lambda}_{k,j}|$
and measured recovery using the mean absolute percentage error (MAPE), i.e., 
$
\mathrm{MAPE}
=
\frac{100}{KH}
\sum_{k=1}^{K}\sum_{j=1}^{H}
\frac{|\widehat{\tau}_{k,j}-\tau_{k,j}|}{\tau_{k,j}}$.
We match inferred regimes and modes to the ground truth
before computing errors and report median MAPE across datasets.

Results are reported in Fig.~\ref{fig:synthetic}. Both state-space methods can accurately recover the stationary
$20$- and $100$-ms timescales across multiple datasets
(Fig.~\ref{fig:synthetic}A).
ACF-exp and aABC provide two-component approximations of
the observed temporal decay.
In the recovered aABC solution, the posterior medians
underestimated the true timescales, but the $95\%$ credible
intervals contained both ground-truth values. Switching Poisson experiments expose a distinction between
trajectory reconstruction and dynamical identification
(Fig.~\ref{fig:synthetic}B,C). In this case,
both state-space methods achieve held-out latent $R^2=0.82$ in the
illustrative experiment, yet MTS-SLDS enables a higher
balanced regime accuracy ($0.95$ versus $0.54$) and lower
timescale error ($11\%$ versus $44\%$).
This finding reinforces the point that comparable trajectory reconstruction may not lead to accurate recovery of timescale or regimes in switching models; MTS-SLDS successfully addresses this by the proposed initialization and inference framework.

Finally, we assess robustness under switching dynamics ($K=2$)
using multiple MTS-SLDS and SLDS fits per noise and spike count level for Gaussian and Poisson emissions respectively.
Under Gaussian observations, both methods maintain median
timescale errors of approximately $6\%$ across signal-to-noise
ratios (SNR) of $0.5$-$5$ (Fig.~\ref{fig:synthetic}D).
Under Poisson observations, estimate errors increase as counts decrease, but MTS-SLDS achieve $1.7$-$3.3$-fold lower median error
across $0.05$-$0.5$ counts per bin per channel (Fig.~\ref{fig:synthetic}E).
These two methods therefore show comparable robustness to
Gaussian noise, while MTS-SLDS outperform SLDS in Poisson observations and preserve more accurate regime-specific timescales under sparse Poisson observations.  Additionaly, we also evaluate how MTS-SLDS can recovery both decaying timescales and oscillatory frequencies by \eqref{eq:mts_slds_timescale} (see Appendix \ref{app:oscillatory}).

\section{Neural data applications and analyses}
\label{sec:neural_data}
\subsection{Interpretable temporal structure in V4 population activity}
\label{sec:v4_results}

\begin{figure}[htb]
    \centering
\includegraphics[width=0.6\linewidth]{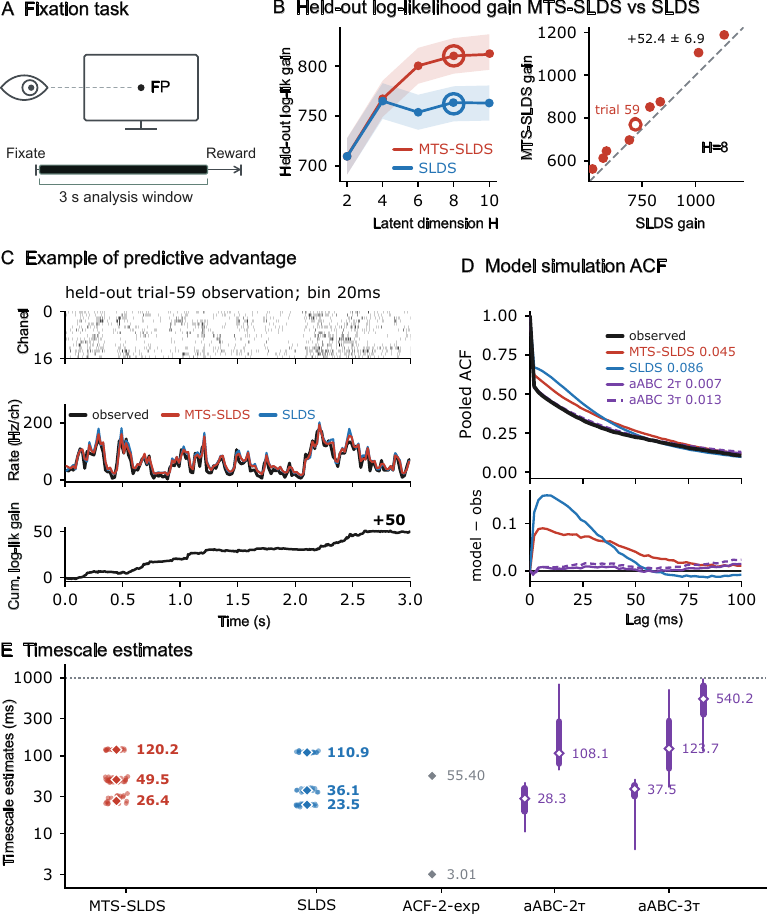}
    \caption{\textbf{Multiple timescale structure of V4
activity during fixation.}
\textbf{A} Fixation task and data analysis schematic.
\textbf{B} Predictive gains over a constant-rate Poisson
baseline: cross-validation (left, mean $\pm$ SE) and
paired test trials at $H=8$ (right).
\textbf{C} Example trial spike train, one-step rate predictions,
and cumulative log-likelihood difference between MTS-SLDS and SLDS.
\textbf{D} Observed and model-generated pooled autocorrelations
and residuals over $100$ ms fitting window, where values in legend denote RMSE of each method.
\textbf{E} Timescale estimates of different methods.
For state-space models, small markers show 5-fold
estimates and diamonds show medians. ACF-exp and aABC fits show estimates from pooled population autocorrelation, and
aABC displays posterior medians and $50\%$/$95\%$ credible
intervals.}
\label{fig:v4}
\end{figure}

We first examine whether  the inferred temporal structure of MTS-SLDS
provides a reproducible and predictive description of
neural population activity.
We analyze a 16-channel  visual area 4 (V4) spike recordings from a macaque doing simple fixation task \cite{engel2016selective,Gieselmann2023}. We focus on the 3-s sustained fixation period and use 2-ms spike count
bins (Fig.~\ref{fig:v4}A). We fit a single regime ($K=1$) state-space model with Poisson emission on the data to characterize dynamics within this sustained behavioral epoch.
We evaluate the model with different latent dimensionality $H\in\{2,4,6,8,10\}$ using five-fold cross-validation and multiple seeds (see details in Appendix~\ref{sec:v4_appendix}.
We find that both methods' prediction saturate at around $H=8$ and
increasing dimension yields little additional
predictive gain; thus we focus subsequent analyses on $H=8$. 

MTS-SLDS provides higher predictive likelihood on all held-out trials, improving log-likelihood by $52.4\pm6.9$ (mean $\pm$ SE) per trial. The example trial illustrates how modest differences in one-step spike rate predictions accumulate into a likelihood advantage of MTS-SLDS over SLDS (Fig.~\ref{fig:v4}C and Fig. \ref{fig:v4_channel_predictions}). In addition, this predictive advantage accompanies better reproduction of the observed temporal correlations. When simulating spike counts from each fitted model and applying the same
pooled-autocorrelation estimation to simulated and recorded counts, we can see MTS-SLDS reproduces the observed autocorrelation more closely than SLDS (Fig.~\ref{fig:v4}D). Both methods yielded three reproducible timescales
(Fig.~\ref{fig:v4}E). MTS-SLDS's inferred timescales center on $26.4$, $49.5$, and
$120.2$~ms, while SLDS produce $23.5$, $36.1$, and $110.9$~ms. This agreement supports a multiscale description of
population dynamics. On the other hand, autocorrelation-based methods provide coarser descriptions due to pooled activity (see Appendix \ref{app:pooling} for pooling effect).
ACF-exp leads to timescales of $3.0$ and $55.4$~ms, while two-component aABC leads to posterior medians of $28.3$ and $108.1$~ms.
Adding a third aABC component shifted the estimates and introduced a poorly constrained component near $540$~ms. The fitted three-component model also matched the pooled
autocorrelation less closely than the two-component model, which implies the two-component fit therefore provided a more parsimonious approximation of this pooled statistic. One possible interpretation is that its fast component summarizes contributions from the two faster latent timescale bands as a single effective decay. Such an effective two-component description can coexist with a richer latent spectrum, highlighting the additional
temporal structure accessible through multichannel modeling with state-space models (see more details in Appendix \ref{sec:area2_appendix}).

\begin{figure}[htb!]
    \centering
    \includegraphics[width=0.6\linewidth]{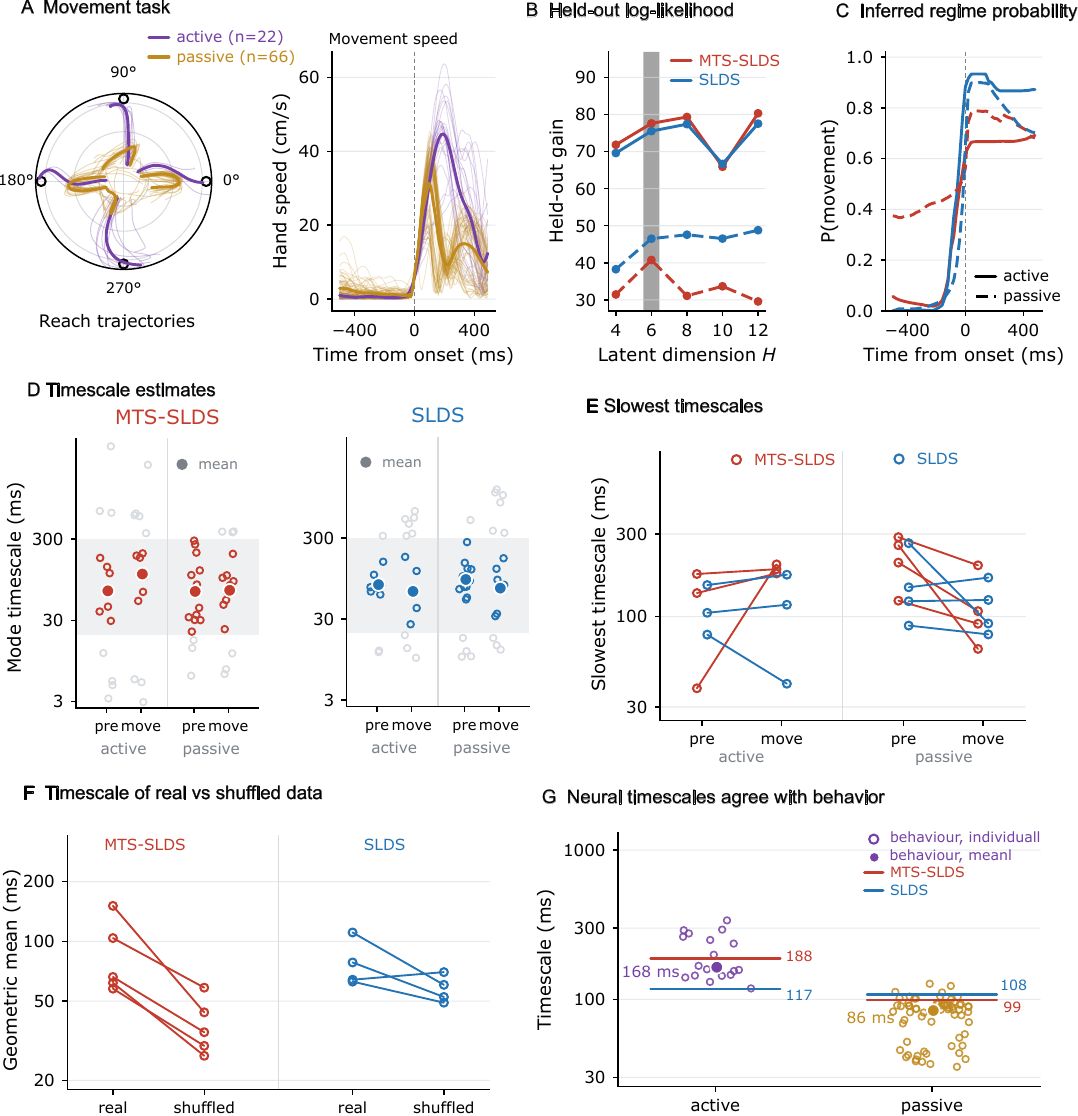}
    \caption{\textbf{Task-related timescales in somatosensory Area 2.}
\textbf{A} Reach trajectories and onset-aligned movement speed during
active reaching (purple) and passive perturbations (orange).
\textbf{B} Held-out predictive log-likelihood gain versus latent
dimension, where gray shading marks the selected $H=6$.
\textbf{C} Inferred movement-regime probabilities with MTS-SLDS and SLDS.
\textbf{D} Latent timescale spectra for pre-movement and
movement regimes in the 20-300-ms analysis range.
\textbf{E} Slowest effective timescales across regimes demonstrate different patterns between MTS-SLDS and SLDS.
\textbf{F} Geometric mean of timescales for real and
 time-shuffled recordings. Lines in \textbf{E,F} connect paired estimates.
\textbf{G} Behavioral decay times (individual estimates and means)
compared with median slowest movement-regime timescales. MTS-SLDS inferred timescales are in close agreement with behavior decay times.}
\label{fig:area2_main}
    \label{fig:area2_main}
\end{figure}

\subsection{Task-related timescales in Somatosensory Cortex Area 2}
\label{sec:area2_main}
We next apply the framework to analyze a neural dataset from a macaque performing reaches, with neural activity recorded from somatosensory cortex Area 2 \citep{DANDI_Area2,Lawlor2018}.
The experiment contains two trial types: active
reaching (4 directions) and passive arm perturbations (Fig.~\ref{fig:area2_main}A), where the hand speed is very different during pre-movement and movement epochs in both conditions. We focus on the $(-500,+500)$ ms period around movement onset.
We fit Poisson MTS-SLDS and SLDS separately to each condition, using
20-ms bins and two regimes $K=2$ (see details in Appendix \ref{sec:area2_appendix}.
We select a common latent dimension of $H=6$ from held-out predictive
likelihoods (Fig.~\ref{fig:area2_main}B) and analyze regimes and associated timescales.
Both methods can identify comparable onset-associated changes in regime probability
(Fig.~\ref{fig:area2_main}C and Fig.\ref{fig:area2_diagnostics}).

We then examine the fitted timescale spectra (Fig.~\ref{fig:area2_main}D) and compared
regimes using the slowest mode (Fig.~\ref{fig:area2_main}E) within a 20-300-ms analysis band in consideration of bin size and trial time window.
For the movement-associated regime, MTS-SLDS identify distinct median timescales
 during active reaching and during passive perturbation (188~ms vs. 99~ms), which is
close to the corresponding behavioral decay times of 168 and 86~ms estimated by ACF fit. However, SLDS provides similar timescales, 117 and 108~ms between the two trial types (Fig.~\ref{fig:area2_main}G).
This agreement suggests that MTS-SLDS provides more interpretable timescales inference corresponding to behavior data and underlying physiological principles. In addition, we perform a shuffle analysis: we permute time bins within trials; both methods tend to find much smaller timescales (Fig.~\ref{fig:area2_main}F) underlying this shuffled data, which supports the fidelity of the learned timescales from data.
A separate analysis using the sPCF-EM method 
\citep{song2023unsupervised} also yields no clear distinction of timescales and no relation with the behavioral timescale ordering (Fig.~\ref{fig:area2_spcf}), which further supports MTS-SLDS as an effective approach for regime-specific timescale estimates. 
Appendix~\ref{sec:area2_appendix} details these analyses.

\section{Discussion}
\label{sec:discussion}
Here, we introduce MTS-SLDS, a framework for recovering regime-specific latent timescales from neural population recordings. It combines a multi-lag moment initialization, which constrains decay rates using temporal dependencies across several observation lags, with regime-conditioned Laplace EM, which keeps the dynamical statistics of each regime separate during learning. The result is a fitted switching model whose transition eigenvalues can be read as timescales: in simulations, MTS-SLDS recovers the prescribed spectra and switching structure across noise levels and spike counts where a standard SLDS does not, and in neural recordings it yields regime-specific timescales that can be compared across behavioral states.
Ablations (Appendix~\ref{app:ablation}) indicate that both components contribute: regime-conditioned inference recovers much of the regime information from the baseline's own parameters, and multi-lag initialization accounts for the remaining gap in timescale error.


We note several limitations to this model. MTS-SLDS estimates \emph{effective} timescales under the assumed model and cannot separate intrinsic circuit dynamics from temporally structured inputs \citep{chaudhuri2015large,vanmeegen2021microscopic}; a fitted transition matrix may absorb persistent drive from unobserved regions or behavioral variables. Weakly observed modes, closely spaced decay rates, short regime visits, and limited recording duration can restrict recoverability. Multi-lag initialization relies on approximate stationarity within candidate regimes, while regime-conditioned learning still uses Gaussian, Laplace, and mixture-reduction approximations. Neither component guarantees globally optimal or unbiased estimates. Future work should quantify uncertainty in the recovered spectrum and its sensitivity to latent dimension, regime number, and observation model.



\section{Statements}
\subsection*{AI use statement}

In this work, we used generative AI tools to assist interpreting results.
We have not used generative AI tools for other tasks with required disclosure.
Additionally, we used generative AI tools to assist code programming and language polishing. We have reviewed all AI-assisted work. LLM-generated code and sentences have been verified for correctness
by authors. We take responsibility for the final content of this work,
including text, claims or artifacts produced with the aid of generative AI.

\subsection*{Ethics Statement}
Here, we aim to make methodological and neuroscientific insights, and do not note any negative societal or ethical implications.

\subsection*{Reproducibility Statement}
Our work can be reproduced in a straightforward way. The neural datasets are provided in \citep{DANDI_Area2} and \citep{Gieselmann2023}, with all pre-processing techniques detailed in \citep{Lawlor2018}, \citep{engel2016selective}, and in the Appendix of this work. Moreover, the code is provided as a supplementary file for review and  will be put online upon acceptance. 



\bibliography{iclr2027_conference}
\bibliographystyle{iclr2027_conference}

\appendix

\section{Latent Timescales and Observed Correlations}
\label{app:timescale_theory}

We derive the fixed-regime covariance identities underlying the timescale
interpretation and multi-lag initialization, then distinguish them from
correlations in a switching process. Observation parameters remain
regime-dependent, as in Section~\ref{sec:mts_slds_model}.

\subsection{Fixed-regime dynamics and Gaussian observations}
\label{app:gaussian_timescales}

Hold regime $k$ fixed, assume $\rho(A_k)<1$, and initialize the
Gaussian latent process in its stationary distribution. Its mean and
covariance satisfy
\begin{equation}
\bar x_k=(I-A_k)^{-1}b_k,
\qquad
\Sigma_k=A_k\Sigma_kA_k^\top+Q_k.
\label{eq:mts_slds_stationary_moments}
\end{equation}
Define the future-first covariance
$\Gamma_{y,k}(\ell)=\operatorname{Cov}(y_{t+\ell},y_t)$.
With independent, temporally white observation noise,
\begin{align}
\Gamma_{x,k}(\ell)
&=\operatorname{Cov}(x_{t+\ell},x_t)=A_k^\ell\Sigma_k,\nonumber\\
\Gamma_{y,k}(\ell)
&=C_kA_k^\ell\Sigma_kC_k^\top
  +\mathbf1_{\{\ell=0\}}R_k,
\qquad \ell=0,1,\ldots.
\label{eq:mts_slds_gaussian_lag}
\end{align}
This follows by iterating the centered dynamics: future innovations
are independent of $x_t$. Observation noise contributes only at zero lag.

For diagonalizable $A_k$, let $v_{k,j}$ and $w_{k,j}^\top$ be
right eigenvectors and dual left rows, with
$w_{k,j}^\top v_{k,r}=\delta_{jr}$. Then
\begin{equation}
\Gamma_{y,k}(\ell)
=\sum_{j=1}^H\lambda_{k,j}^\ell
(C_kv_{k,j})(w_{k,j}^\top\Sigma_kC_k^\top),
\qquad \ell>0.
\label{eq:mts_slds_modal_covariance}
\end{equation}
Writing a decaying eigenvalue as $\lambda=re^{i\theta}$,
with $0<r<1$ and principal phase $\theta=\arg\lambda$, gives
\begin{equation}
\lambda^\ell=e^{-\ell\Delta t/\tau}e^{i\ell\theta},
\qquad
\tau=-\frac{\Delta t}{\log r},
\qquad
f=\frac{|\theta|}{2\pi\Delta t}.
\label{eq:mts_slds_modal_decay}
\end{equation}
Positive real eigenvalues describe nonoscillatory decay; negative
real eigenvalues alternate sign. Complex-conjugate pairs describe
real damped oscillations with a shared $\tau$ and $f$. The frequency
is determined by phase, not the imaginary part of $\lambda$ alone,
and is subject to sampling aliasing.

For a channel with positive variance,
\begin{align}
\rho_{n,k}(\ell)
&=\sum_j\alpha_{n,k,j}\lambda_{k,j}^\ell,
\qquad \ell>0,\nonumber\\
\alpha_{n,k,j}
&=\frac{(c_{n,k}^\top v_{k,j})
(w_{k,j}^\top\Sigma_kc_{n,k})}
{c_{n,k}^\top\Sigma_kc_{n,k}+[R_k]_{nn}}.
\label{eq:mts_slds_modal_acf}
\end{align}
These coefficients may vanish, be signed, or combine in conjugate
pairs: an observed ACF need not expose every latent mode or form a
nonnegative exponential mixture. Defective matrices can introduce
polynomial prefactors, and nonnormal dynamics can exhibit transient
amplification. Thus $\tau$ characterizes modal exponential decay,
not the entire finite-time response. We report finite positive
relaxation times only for $0<|\lambda|<1$.

\subsection{Poisson observations and moment conversion}
\label{app:poisson_timescales}

For the same stationary Gaussian latent process, use the exponential
Poisson emissions in Eq.~\eqref{eq:mts_slds_poisson_obs}, with counts
conditionally independent across channel-time indices. The mean count is
\begin{equation}
\mu_{n,k}
=\exp\!\left(d_{n,k}+c_{n,k}^\top\bar x_k
+\tfrac12c_{n,k}^\top\Sigma_kc_{n,k}\right).
\label{eq:mts_slds_poisson_mean}
\end{equation}
The Gaussian identity
$\mathbb E[e^{a+b}]=\mathbb E[e^a]\mathbb E[e^b]
e^{\operatorname{Cov}(a,b)}$
and conditional Poisson variance give
\begin{equation}
\begin{split}
[\Gamma_{y,k}(\ell)]_{nm}
={}&\mu_{n,k}\mu_{m,k}
\left[\exp\!\left(c_{n,k}^\top A_k^\ell\Sigma_kc_{m,k}\right)-1\right]\\
&+\mathbf1_{\{\ell=0\}}\delta_{nm}\mu_{n,k},
\qquad \ell\geq0.
\end{split}
\label{eq:mts_slds_poisson_covariance}
\end{equation}
Consequently, the elementwise conversion
\begin{align}
[M_{\ell,k}]_{nm}
&=\log\!\left(1+
\frac{[\Gamma_{y,k}(\ell)]_{nm}
-\mathbf1_{\{\ell=0\}}\delta_{nm}\mu_{n,k}}
{\mu_{n,k}\mu_{m,k}}\right),\nonumber\\
M_{\ell,k}&=C_kA_k^\ell\Sigma_kC_k^\top,
\qquad \ell\geq0,
\label{eq:mts_slds_poisson_moment_conversion}
\end{align}
recovers the log-intensity covariance \citep{buesing2012spectral}.
At positive lags, no Poisson diagonal subtraction is needed.

Raw count correlations are nonlinear functions of latent covariance.
For example, a scalar log-intensity covariance $v a^\ell$, with $v>0$ and
$0<a<1$, produces a count covariance proportional to
$\sum_{p\geq1}(v^p/p!)a^{p\ell}$, introducing decay times $\tau/p$.
Direct exponential fits to count ACFs therefore need not isolate the
original latent timescales. The conversion requires jointly Gaussian
latent states and an exponential link; it is not exact for arbitrary
switching mixtures or other observation links.

\subsection{Switching and coordinate invariance}
\label{app:switching_timescales}

In an SLDS, $\operatorname{Cov}(x_t\mid z_t=k)$ generally differs
from the fixed-regime $\Sigma_k$ because it retains preceding-regime
history. Even a switch-free interval can begin outside stationarity.
Along a prescribed regime path, perturbations with identical innovations
satisfy
\begin{equation}
\delta x_{t+\ell}=\Phi_{t,\ell}\delta x_t,
\qquad
\Phi_{t,\ell}=A_{z_{t+\ell}}\cdots A_{z_{t+1}}.
\label{eq:mts_slds_switching_propagator}
\end{equation}
Observed correlations consequently depend on regime paths, means,
and emission mappings, not only on individual spectra. The mean
visit duration $\Delta t/(1-P_{kk})$, for $P_{kk}<1$, is distinct
from a regime's relaxation times. Stability of each $A_k$ does not
by itself ensure stability of the switching process.

A common invertible transformation $x_t'=Sx_t$ gives
\begin{equation}
A_k'=SA_kS^{-1},\qquad
C_k'=C_kS^{-1},\qquad
Q_k'=SQ_kS^\top,
\label{eq:mts_slds_coordinate_invariance}
\end{equation}
with $b_k'=Sb_k$ and correspondingly transformed initial moments.
Similarity preserves timescales and frequencies. Separate transformations
$S_k$ instead yield $S_kA_kS_i^{-1}$ on a transition $i\to k$,
which generally falls outside the destination-only transition
parameterization. The initializer must therefore specify the relative
latent coordinates across regimes
(Appendix~\ref{app:moment_handoff}).

These identities describe spectral interpretation, not guaranteed
recovery: weakly observed modes, regime uncertainty, and misspecification
can limit estimation even when reconstruction is accurate.

\section{Multi-Lag Moment Initialization}
\label{app:moment_initialization}

The initializer constructs a pooled temporal representation, obtains
provisional regime labels, and estimates dynamics from regime-local
moments. All estimates use training trials; lag pairs never cross
trial boundaries. The stationary identities motivate this procedure
but need not hold exactly under estimated segmentation or switching
transients.

\subsection{Empirical moments and observation-model conversion}
\label{app:moment_empirical}

For a sample group $g$, let $\widehat\mu_g$ be its observation mean
and $\mathcal I_{\ell,g}$ its admissible trial/start-bin pairs.
Estimate
\begin{equation}
\widehat\Gamma_{\ell,g}
=\frac{1}{|\mathcal I_{\ell,g}|}
\sum_{(r,t)\in\mathcal I_{\ell,g}}
(y_{r,t+\ell}-\widehat\mu_g)
(y_{r,t}-\widehat\mu_g)^\top,
\qquad \ell>0.
\label{eq:mts_slds_empirical_lags}
\end{equation}
Groups initially pool training data and later contain one assigned
regime. Writing $\widetilde\Gamma_{\ell,g}$ for the regularized
estimate, set $\widehat M_{\ell,g}=\widetilde\Gamma_{\ell,g}$
for Gaussian observations. For Poisson observations, use
\begin{equation}
[\widehat M_{\ell,g}]_{nm}
=\log\!\left(1+
\frac{[\widetilde\Gamma_{\ell,g}]_{nm}}
{\widehat\mu_{n,g}\widehat\mu_{m,g}}\right).
\label{eq:mts_slds_empirical_count_transform}
\end{equation}
Mean estimates and logarithm arguments must be positive; negative
covariances themselves are admissible. Finite-sample regularization
can alter the population identity
\begin{equation}
M_{\ell,k}=C_kA_k^\ell\Sigma_kC_k^\top,
\qquad \ell>0.
\label{eq:mts_slds_moment_reference}
\end{equation}
At zero lag, Poisson moments require subtraction of
$\operatorname{diag}(\mu_k)$ before conversion. Gaussian zero-lag
moments instead require separation of $C_k\Sigma_kC_k^\top$ and $R_k$.

\subsection{Windowed temporal segmentation}
\label{app:moment_switching}

Construct $U\in\mathbb R^{D\times H}$, $U^\top U=I$, from the
leading left singular vectors of concatenated pooled positive-lag
moments and their transposes. Compare projected signal moments with
time-shuffled null moments, select the longest consecutive lag band
above the threshold, and refine the projection and band. Under switching,
this is a feature construction, not an exact rank-$H$ realization.

Using the pooled mean $\widehat\mu$, form
\begin{equation}
\widetilde x_{r,t}=
\begin{cases}
U^\top\operatorname{diag}(\widehat\mu)^{-1}
(y_{r,t}-\widehat\mu), & \text{Poisson},\\
U^\top(y_{r,t}-\widehat\mu), & \text{Gaussian}.
\end{cases}
\label{eq:mts_slds_segmentation_projection}
\end{equation}
These are segmentation features, not posterior latent estimates.
For window $w$, estimate their lag covariance $\widehat K_w(\ell)$
and define
\begin{equation}
r_w(\ell)=\frac{\operatorname{tr}\widehat K_w(\ell)}
{\operatorname{tr}\widehat K_w(0)},
\qquad
\phi_w(\ell)=\log\max\{r_w(\ell),\epsilon_{\mathrm{acf}}\},
\quad \ell\in\mathcal L_{\mathrm{desc}}.
\label{eq:mts_slds_window_descriptor}
\end{equation}
The positive floor discards negative ACF values; this descriptor
summarizes aggregate decay rather than resolving individual modes
or fully representing oscillatory structure. Signed moment matrices
are still used in the transition regression below.

Standardize descriptors, apply $K$-means, and obtain cluster Gaussian
log scores $a_w(k)$. A sticky Viterbi pass assigns window labels:
\begin{equation}
\widehat z^{\,\mathrm{win}}_{1:J}
=\arg\max_{z_{1:J}}\left[
\sum_{w=1}^J a_w(z_w)
+\beta_{\mathrm{seg}}\sum_{w=2}^J\mathbf1\{z_w=z_{w-1}\}
\right].
\label{eq:mts_slds_segmentation_viterbi}
\end{equation}
Overlapping windows vote on bin labels. Include terminal windows and
assign any uncovered bins the nearest covered label within the trial.

\subsection{Regime-local multi-lag regression}
\label{app:moment_regression}

For regime $k$, use a common set of starts satisfying
$\widehat z_{r,t:t+L_{\max}}=k$ and $t+L_{\max}\leq T_r$,
where $T_r$ is the trial length. Estimate $\widehat\mu_k$ from all
bins assigned to $k$, then recompute moments from the original
observations using these starts at every lag through $L_{\max}$.
Repeat the null-moment comparison within each regime and set
$S_{\ell,k}=U^\top\widehat M_{\ell,k}U$.

At the population level, if $\mathsf G_k=U^\top C_k$ is invertible,
\begin{equation}
S_{\ell+1,k}=A_{U,k}S_{\ell,k},
\qquad
A_{U,k}=\mathsf G_kA_k\mathsf G_k^{-1}.
\label{eq:mts_slds_projected_shift}
\end{equation}
Thus the projected shift has the same eigenvalues as $A_k$.
Unique unregularized recovery additionally requires the horizontal
stack of fitted $S_{\ell,k}$ to have row rank $H$; ridge regularization
does not establish identifiability.

Let $\mathsf D_k=\mathsf L_k\mathsf L_k^\top\succ0$ be the
regularized symmetric lag-one reference used for coordinate
preconditioning. The unregularized matrix
$\tfrac12(S_{1,k}+S_{1,k}^\top)$ need not be positive definite
and is not a zero-lag covariance.
Set $\overline S_{\ell,k}=\mathsf L_k^{-1}S_{\ell,k}\mathsf L_k^{-\top}$
and choose adjacent-lag indices
$\mathcal B_k\subseteq\{1,\ldots,L_{\max}-1\}$. Solve
\begin{equation}
\min_B\sum_{\ell\in\mathcal B_k}
\|\overline S_{\ell+1,k}-B\overline S_{\ell,k}\|_F^2
+\lambda_{\mathrm{ridge}}\|B\|_F^2.
\label{eq:mts_slds_multilag_objective}
\end{equation}
For an invertible regularized normal matrix,
\begin{align}
\widehat A_{n,k}
&=\left[\sum_{\ell\in\mathcal B_k}
\overline S_{\ell+1,k}\overline S_{\ell,k}^\top\right]
\left[\sum_{\ell\in\mathcal B_k}
\overline S_{\ell,k}\overline S_{\ell,k}^\top
+\lambda_{\mathrm{ridge}}I\right]^{-1},
\label{eq:mts_slds_spectral_realization}\\
\widehat A_{U,k}&=\mathsf L_k\widehat A_{n,k}\mathsf L_k^{-1}.
\label{eq:mts_slds_projected_dynamics_init}
\end{align}
This fits multiple covariance-shift relations, not a regression of
$\widetilde x_t$ on its history. Preconditioning changes weighting;
it does not remove errors in the empirical moments. Regimes with
insufficient admissible starts use the pooled operator before
coordinate conversion.

\subsection{Emission parameters, coordinates, and covariance initialization}
\label{app:moment_handoff}

Estimate the zero-lag signal matrix
$\widehat M_{0,k}^{\,\mathrm{sig}}\approx C_k\Sigma_kC_k^\top$
and choose rank-$H$ emission factors $C_k^{(0)}$ with an explicit
relative coordinate convention across regimes. If
$\mathsf G_k=U^\top C_k^{(0)}$ is nonsingular, set
\begin{equation}
\begin{split}
A_k^{(0)}&=\mathsf G_k^{-1}\widehat A_{U,k}\mathsf G_k,\\
\Sigma_k^{(0)}&=\mathsf G_k^{-1}
(U^\top\widehat M_{0,k}^{\,\mathrm{sig}}U)\mathsf G_k^{-\top}.
\end{split}
\label{eq:mts_slds_coordinate_handoff}
\end{equation}
Within-regime moments determine spectra but not relative latent
coordinates across regimes; independent factorizations do not resolve
this ambiguity.

Choose zero latent reference means and $b_k^{(0)}=0$, with
\begin{equation}
d_k^{(0)}=
\begin{cases}
\log\widehat\mu_k-\tfrac12\operatorname{diag}
(C_k^{(0)}\Sigma_k^{(0)}(C_k^{(0)})^\top), & \text{Poisson},\\
\widehat\mu_k, & \text{Gaussian}.
\end{cases}
\label{eq:mts_slds_emission_offset_init}
\end{equation}
Choose Gaussian $R_k^{(0)}\succ0$ consistently with the zero-lag
decomposition. A stationary reference covariance
$\Sigma_{\mathrm{ref},k}$ is compatible with the candidate
\begin{equation}
Q_k^{(0)}=\Sigma_{\mathrm{ref},k}
-A_k^{(0)}\Sigma_{\mathrm{ref},k}(A_k^{(0)})^\top
\label{eq:mts_slds_process_covariance_init}
\end{equation}
when the result is positive definite. Matching the preceding emission
reference requires $\Sigma_{\mathrm{ref},k}=\Sigma_k^{(0)}$;
other choices or positive-definite corrections need not preserve that
reference. For $\Sigma_{\mathrm{ref},k}=I$, positive definiteness
requires $\|A_k^{(0)}\|_2<1$, not merely $\rho(A_k^{(0)})<1$.

For adjacent-label counts $N_{ik}$, a row-stochastic persistence
reference $P_{\mathrm{prior}}$, and pseudocount mass $\eta>0$, use
\begin{equation}
P_{ik}^{(0)}=\frac{N_{ik}+\eta[P_{\mathrm{prior}}]_{ik}}
{\sum_jN_{ij}+\eta}.
\label{eq:mts_slds_transition_init}
\end{equation}
Initialize $\pi_k^{(0)}=1/K$, $\mu_0^{(0)}=0$, and $V_0^{(0)}=I$
in the selected coordinates, then refine with RC-L-EM.

\section{Regime-Conditioned Inference and Learning}
\label{app:rclem_details}

Each iteration holds parameters fixed for forward filtering and backward
smoothing, then updates the selected parameter blocks. We suppress trial
indices; sufficient statistics sum over independent trials without
cross-trial transitions. The regime $z_t=k$ selects the dynamics into
$x_t$ and the emission at $t$. The initial distribution is
$z_1\sim\operatorname{Cat}(\pi)$ and
$x_1\mid z_1=k\sim\mathcal N(\mu_0,V_0)$.

\subsection{Posterior approximation and dynamical statistics}
\label{app:rclem_posterior}

Write $\gamma_t^f(k)$ and $\gamma_t^s(k)$ for approximate filtered
and smoothed regime probabilities, and $\xi_t(i,k)$ for approximate
smoothed adjacent-regime probabilities. Variational Laplace EM in
\texttt{ssm} alternates $q_z(z_{1:T})$ and a global Gaussian Laplace
approximation $q_x(x_{1:T})$ \citep{zoltowski2020general}; its
factorization makes the continuous-state approximation independent
of the candidate regime. RC-L-EM instead retains regime- and
pair-conditioned Gaussian moments. For $f_t=f(x_{t-1},x_t)$,
\begin{align}
\mathcal M_{t,k}^{\mathrm{MF}}(f)
&=\gamma_t^s(k)\mathbb E_{q_x}[f_t],\nonumber\\
\mathcal M_{t,k}^{\mathrm{RC}}(f)
&=\sum_i\xi_t(i,k)\mathbb E_{q_t^{ik}}[f_t].
\label{eq:rclem_moment_comparison}
\end{align}
Each expression uses its own method's responsibilities. The regression
algebra is shared; its conditional state moments differ
(Appendix~\ref{app:ssm_baseline}).

\subsection{Forward filtering}
\label{app:rclem_filtering}

Represent the filtered state by
\begin{equation}
q_f(x_t,z_t=k)=\gamma_t^f(k)
\mathcal N(x_t;m_t^{(k)},V_t^{(k)}).
\label{eq:rclem_filter_joint}
\end{equation}
For $t\geq2$, compute
\begin{equation}
\bar\gamma_t(k)=\sum_i\gamma_{t-1}^f(i)P_{ik},
\qquad
\omega_{i\to k,t}=\frac{\gamma_{t-1}^f(i)P_{ik}}{\bar\gamma_t(k)}.
\label{eq:rclem_mixing_weights}
\end{equation}
For positive-mass destinations, moment-match the incoming mixture:
\begin{align}
\bar m_{t-1}^{(k)}&=\sum_i\omega_{i\to k,t}m_{t-1}^{(i)},\nonumber\\
\bar V_{t-1}^{(k)}&=\sum_i\omega_{i\to k,t}
\left[V_{t-1}^{(i)}+\delta_{i,k,t}\delta_{i,k,t}^\top\right],
\label{eq:rclem_mixed_moments}
\end{align}
where $\delta_{i,k,t}=m_{t-1}^{(i)}-\bar m_{t-1}^{(k)}$.
The state prediction is
\begin{equation}
a_t^{(k)}=A_k\bar m_{t-1}^{(k)}+b_k,
\qquad
S_t^{(k)}=A_k\bar V_{t-1}^{(k)}A_k^\top+Q_k.
\label{eq:rclem_prediction}
\end{equation}
At $t=1$, use $(\bar\gamma_1(k),a_1^{(k)},S_1^{(k)})
=(\pi_k,\mu_0,V_0)$. Evaluate the predictive evidence
\begin{equation}
\mathcal L_t(k)=\int p_\theta(y_t\mid x,z_t=k)
\mathcal N(x;a_t^{(k)},S_t^{(k)})\,dx.
\label{eq:rclem_predictive_evidence}
\end{equation}
Its approximation to the model's conditional evidence already reflects
collapse of the incoming mixture.

For Gaussian observations, let
$e_t^{(k)}=y_t-C_ka_t^{(k)}-d_k$ and
$\Omega_t^{(k)}=C_kS_t^{(k)}C_k^\top+R_k$. Then
\begin{align}
\mathcal L_t(k)&=\mathcal N(y_t;C_ka_t^{(k)}+d_k,\Omega_t^{(k)}),
\label{eq:rclem_gaussian_evidence}\\
\mathsf K_t^{(k)}&=S_t^{(k)}C_k^\top(\Omega_t^{(k)})^{-1},\nonumber\\
m_t^{(k)}&=a_t^{(k)}+\mathsf K_t^{(k)}e_t^{(k)},\nonumber\\
V_t^{(k)}&=S_t^{(k)}-\mathsf K_t^{(k)}\Omega_t^{(k)}
(\mathsf K_t^{(k)})^\top.
\label{eq:rclem_gaussian_update}
\end{align}
These updates are exact for the Gaussian prediction, not generally
for the uncollapsed SLDS posterior.

For Poisson observations, define the local log joint
\begin{equation}
\begin{split}
h_{t,k}(x)={}&\log\mathcal N(x;a_t^{(k)},S_t^{(k)})
+y_t^\top(C_kx+d_k)\\
&-\mathbf1^\top e^{C_kx+d_k}-\sum_n\log(y_{nt}!).
\end{split}
\label{eq:rclem_local_density}
\end{equation}
For $S_t^{(k)}\succ0$, this is strictly concave. Set
\begin{align}
m_t^{(k)}&=\arg\max_x h_{t,k}(x),
\qquad V_t^{(k)}=\Lambda_{t,k}^{-1},\nonumber\\
\Lambda_{t,k}&=(S_t^{(k)})^{-1}
+C_k^\top\operatorname{diag}(e^{C_km_t^{(k)}+d_k})C_k,
\label{eq:rclem_local_laplace}
\end{align}
with Laplace evidence
\begin{equation}
\mathcal L_t(k)\simeq
\exp(h_{t,k}(m_t^{(k)}))(2\pi)^{H/2}|\Lambda_{t,k}|^{-1/2}.
\label{eq:rclem_poisson_evidence}
\end{equation}
For either observation model, we normalize in the log domain:
\begin{equation}
\gamma_t^f(k)=\frac{\bar\gamma_t(k)\mathcal L_t(k)}
{\sum_j\bar\gamma_t(j)\mathcal L_t(j)}.
\label{eq:rclem_gamma_update}
\end{equation}
 Gaussian $K=1$ reduces to Kalman
filtering; Poisson $K=1$ still uses local Laplace approximations.

\subsection{Backward smoothing and pair moments}
\label{app:rclem_smoothing}

The backward pass incorporates future observations and supplies
$\gamma_t^s(k)$, $\xi_t(i,k)$, and conditional state moments
\citep{song2022modeling,song2023unsupervised}. Write
$q_t^k=\mathcal N(m_{t\mid T}^{(k)},V_{t\mid T}^{(k)})$
for the single-time approximation, and
\begin{equation}
q_t^{ik}=\mathcal N\!\left(
\begin{bmatrix}x_{t-1}\\x_t\end{bmatrix};
\begin{bmatrix}m_{t,-}^{ik}\\m_{t,+}^{ik}\end{bmatrix},
\begin{bmatrix}V_{t,-}^{ik}&(U_t^{ik})^\top\\
U_t^{ik}&V_{t,+}^{ik}\end{bmatrix}\right)
\label{eq:rclem_pair_approximation}
\end{equation}
for the state approximation conditional on $(z_{t-1},z_t)=(i,k)$.
The cross moment needed for dynamics learning is
\begin{equation}
\mathbb E_{q_t^{ik}}[x_tx_{t-1}^\top]
=U_t^{ik}+m_{t,+}^{ik}(m_{t,-}^{ik})^\top.
\label{eq:rclem_pair_cross_moment}
\end{equation}
Pair-specific predictions retain the predecessor component:
\begin{equation}
a_t^{ik}=A_km_{t-1}^{(i)}+b_k,
\qquad
S_t^{ik}=A_kV_{t-1}^{(i)}A_k^\top+Q_k,
\label{eq:rclem_pair_prediction}
\end{equation}
with prior cross-covariance $V_{t-1}^{(i)}A_k^\top$.
The collapsed destination prediction does not retain this information.

Use pair moments directly for dynamics and single-time moments for
emissions. These local approximations need not define one globally
consistent trajectory posterior; regime weights or products of state
means alone do not supply the required cross-covariances.

\subsection{Dynamics and transition updates}
\label{app:rclem_mstep}

Let $v_t=[x_{t-1}^\top,1]^\top$ and $W_k=[A_k\;b_k]$.
Accumulate
\begin{align}
N_k&=\sum_{t=2}^T\sum_i\xi_t(i,k),\nonumber\\
S_{vv,k}&=\sum_{t=2}^T\sum_i\xi_t(i,k)
\mathbb E_{q_t^{ik}}[v_tv_t^\top],\nonumber\\
S_{xv,k}&=\sum_{t=2}^T\sum_i\xi_t(i,k)
\mathbb E_{q_t^{ik}}[x_tv_t^\top],\nonumber\\
S_{xx,k}&=\sum_{t=2}^T\sum_i\xi_t(i,k)
\mathbb E_{q_t^{ik}}[x_tx_t^\top].
\label{eq:rclem_stats}
\end{align}
For positive occupancy and nonsingular $S_{vv,k}$, unconstrained,
unregularized regression gives
\begin{equation}
W_k^{\mathrm{new}}=S_{xv,k}S_{vv,k}^{-1},
\label{eq:rclem_dynamics_mstep}
\end{equation}
and, evaluated at the updated $W_k$,
\begin{equation}
Q_k^{\mathrm{new}}=\frac{S_{xx,k}-W_kS_{xv,k}^\top
-S_{xv,k}W_k^\top+W_kS_{vv,k}W_k^\top}{N_k}.
\label{eq:rclem_constrained_covariance}
\end{equation}
Use linear solves. Priors, covariance restrictions, and stability
constraints modify these updates; a real Schur representation alone
does not enforce $\rho(A_k)<1$.

For positive expected departure counts,
\begin{equation}
P_{ik}^{\mathrm{new}}=\frac{\sum_{t=2}^T\xi_t(i,k)}
{\sum_{t=2}^T\sum_j\xi_t(i,j)}.
\label{eq:rclem_transition_mstep}
\end{equation}
An interior MAP update with Dirichlet concentrations $\alpha_{ik}$
adds $\alpha_{ik}-1$ to expected counts when all adjusted counts are
positive. A sticky prior increases the diagonal concentration;
posterior-mean smoothing is a different update.

\subsection{Observation parameters and fixed parameter blocks}
\label{app:rclem_observation_updates}

Emission updates use $\gamma_t^s(k)$ and $q_t^k$, including $t=1$.
For Gaussian observations, let $w_t=[x_t^\top,1]^\top$,
$F_k=[C_k\;d_k]$, $N_k^y=\sum_t\gamma_t^s(k)$, and
\[
\begin{aligned}
G_k&=\sum_t\gamma_t^s(k)\mathbb E[w_tw_t^\top],\qquad
Y_k=\sum_t\gamma_t^s(k)y_ty_t^\top,\\
H_k&=\sum_t\gamma_t^s(k)y_t\mathbb E[w_t]^\top,
\end{aligned}
\]
where expectations are under $q_t^k$. The unregularized updates are
\begin{equation}
F_k^{\mathrm{new}}=H_kG_k^{-1},
\qquad
R_k^{\mathrm{new}}=\frac{Y_k-H_kG_k^{-1}H_k^\top}{N_k^y},
\label{eq:rclem_gaussian_obs_mstep}
\end{equation}
for positive occupancy and nonsingular $G_k$.

For Poisson observations, numerically maximize, for each neuron and
regime, the fixed-moment objective
\begin{equation}
\begin{split}
\mathcal Q_{n,k}(c,d)=\sum_t\gamma_t^s(k)\Big[
&y_{nt}(c^\top m_{t\mid T}^{(k)}+d)\\
&-\exp\!\left(d+c^\top m_{t\mid T}^{(k)}
+\tfrac12c^\top V_{t\mid T}^{(k)}c\right)\Big],
\end{split}
\label{eq:rclem_poisson_obs_mstep}
\end{equation}
where $(c,d)=(c_{n,k},d_{n,k})$; parameter-independent terms are
omitted. The variance term accounts for continuous-state uncertainty.

Fixed parameter blocks retain their initialized values. Rerun inference
after updates and once at the final parameters for posterior summaries.
RC-L-EM is an approximate EM-style procedure: moment matching and
Laplace approximations do not provide a general guarantee of monotonic
marginal-likelihood ascent or improved timescales at each iteration.

\section{Baseline Timescale Estimators and SLDS Inference}
\label{app:timescale_baselines}

This appendix describes ACF-exponential fitting, aABC, and SLDS
inference, and clarifies their relation to autoregressive models.
ACF fitting and aABC compare temporal summaries; SLDS methods fit
multivariate observations through latent dynamics and extract
regime-specific timescales using
Eq.~\eqref{eq:mts_slds_timescale}.

\subsection{Exponential fitting of the autocorrelation}
\label{app:acf_exp}

For a stationary scalar signal with positive variance, define
$\rho(\ell)=\operatorname{Cov}(s_{t+\ell},s_t)/
\operatorname{Var}(s_t)$.
A conventional empirical estimator is
\begin{equation}
\widehat\rho(\ell)=
\frac{\sum_{t=1}^{T-\ell}(s_{t+\ell}-\bar s)(s_t-\bar s)}
{\sum_{t=1}^{T}(s_t-\bar s)^2},
\qquad 1\leq\ell<T,
\label{eq:baseline_empirical_acf}
\end{equation}
where $\bar s$ is the sample mean. For repeated trials, we restrict
lagged products to within-trial pairs. Centering, normalization,
and pooling choices define the empirical summary.

ACF-exponential fitting uses nonlinear least squares to fit $M$
decay components:
\begin{equation}
\min_{\{a_j,\tau_j\},\,b_{\mathrm{acf}}}
\sum_{\ell\in\mathcal L}v_\ell
\left[
\widehat\rho(\ell)-b_{\mathrm{acf}}
-\sum_{j=1}^{M}a_j e^{-\ell\Delta t/\tau_j}
\right]^2,
\qquad \tau_j>0.
\label{eq:baseline_acf_fit}
\end{equation}
The fitting configuration specifies positive lags $\mathcal L$,
nonnegative weights $v_\ell$, amplitude constraints, and whether
to fix $b_{\mathrm{acf}}=0$. Taking logarithms does not linearize
a sum of exponentials or an additive offset. The lag window and
finite-sample distortion of the ACF can affect the estimates
\citep{zeraati2022flexible}; this fit does not explicitly model
oscillations.

\subsection{Relation to autoregressive models}
\label{app:ar_timescales}

Autoregressive (AR) models describe observations through their
history. For centered multivariate observations, VAR($p$) specifies
\begin{equation}
y_t=\sum_{r=1}^{p}F_r y_{t-r}+e_t,
\label{eq:baseline_var}
\end{equation}
where $e_t$ is a white innovation process. Its companion matrix
propagates the stacked observation history; each eigenvalue
$\lambda$ with $0<|\lambda|<1$ gives the decay timescale
$-\Delta t/\log|\lambda|$. Complex-conjugate eigenvalues also
encode oscillations through their phase. AR(1) reduces to
$\rho(\ell)=\phi^\ell$ for $|\phi|<1$
\citep{lutkepohl2005new}.
Thus, AR/VAR models can represent multiple timescales, and
ARMA/VARMA representations can encode the same observable poles
as linear-Gaussian state-space models. Our formulation instead
places autoregressive dynamics in a shared latent process and
explicitly models its Gaussian or Poisson observations. The
relevant distinction is latent versus direct observation
modeling, not whether a method uses eigenvalues.

\subsection{Adaptive approximate Bayesian computation}
\label{app:aabc}

Adaptive approximate Bayesian computation (aABC) estimates
simulator parameters by comparing observed and simulated summaries
\citep{zeraati2022flexible}. A common generator combines independent,
stationary Ornstein--Uhlenbeck components:
\begin{equation}
\begin{aligned}
dX_j(s)&=-\frac{X_j(s)}{\tau_j}\,ds
+\sqrt{\frac{2}{\tau_j}}\,dW_j(s),\\
g(s)&=\mu_g+\sigma_g\sum_{j=1}^{M}\sqrt{w_j}X_j(s),
\end{aligned}
\label{eq:baseline_ou_generator}
\end{equation}
where $\tau_j>0$, the $W_j$ are independent Wiener processes,
$w_j\geq0$, and $\sum_jw_j=1$. This gives
$\operatorname{Cov}(g(s+h),g(s))=
\sigma_g^2\sum_jw_j e^{-|h|/\tau_j}$.
An observation simulator converts $g$ into measurements or counts.

For summary $S$, distance $d$, prior $p(\vartheta)$, and tolerance
$\varepsilon$, the ABC target is
\begin{equation}
p_\varepsilon(\vartheta\mid S(y))
\propto p(\vartheta)
\Pr_\vartheta\!\left[
 d\{S(Y^{\mathrm{sim}}),S(y)\}\leq\varepsilon
\right].
\label{eq:baseline_abc_target}
\end{equation}
aABC adapts its proposal and tolerance using importance-weighted
population Monte Carlo. Matching trial structure, binning, and
summary computation reproduces finite-sample effects under the
simulator, but does not guarantee unbiased estimates under
misspecification. The resulting approximate posterior describes
the simulator's timescales; their correspondence to SLDS modes
depends on the generator and observation model.

\subsection{SLDS with variational Laplace EM}
\label{app:ssm_baseline}

We fit the SLDS baseline with variational Laplace EM in
\texttt{ssm} \citep{zoltowski2020general,ssmsoftware}.
The standard initializer fits an emission model, inverts it to
obtain provisional latent trajectories, and fits an autoregressive
HMM to initialize the switching dynamics. Common Gaussian emission
classes use PCA; corresponding Poisson classes apply the selected
link to clipped counts before PCA. Thus, this initializer already
uses temporal fitting rather than arbitrary random dynamics.

Writing $x=x_{1:T}$ and $z=z_{1:T}$, the method approximates the
posterior by $q_x(x)q_z(z)$ and targets the ELBO
\begin{equation}
\mathcal L(q_x,q_z,\theta)
=\mathbb E_{q_xq_z}[\log p_\theta(y,x,z)]
+\mathcal H(q_x)+\mathcal H(q_z),
\label{eq:baseline_vlem_elbo}
\end{equation}
where $\mathcal H$ denotes entropy. This factorization separates
the continuous and discrete trajectories, not individual time
points. For fixed $q_x$, the coordinate-optimal discrete update is
\begin{equation}
q_z^\star(z)\propto
\exp\!\left(\mathbb E_{q_x}[\log p_\theta(y,x,z)]\right).
\label{eq:baseline_vlem_discrete}
\end{equation}
Forward--backward computes the regime and transition marginals
from the expected log factors. Regime-dependent emissions
contribute to these factors; shared emissions cancel from the
relative regime weights. The standard implementation approximates
these expectations by Monte Carlo.

For fixed $q_z$, define
$h(x)=\mathbb E_{q_z}[\log p_\theta(y,x,z)]$ and use the global
trajectory Laplace update
\begin{equation}
\widehat x=\arg\max_x h(x),\qquad
q_x(x)=\mathcal N\!\left(
 x;\widehat x,[-\nabla^2h(\widehat x)]^{-1}
\right).
\label{eq:baseline_vlem_continuous}
\end{equation}
The precision matrix is block-tridiagonal. For our Gaussian SLDS
with state-independent regime transitions, $h$ is quadratic, so
this Gaussian update exactly solves the $q_x$ coordinate step;
the joint factorization remains approximate. For Poisson emissions,
the Gaussian update is an approximation. In general, Laplace
updates need not increase the ELBO \citep{zoltowski2020general}.

Parameter updates use the resulting approximate expectations.
The implementation supports analytic dynamics moments for selected
first-order autoregressive classes and samples other expectations
\citep{ssmsoftware}. Its dynamics update uses
$q_z(z_t=k)\mathbb E_{q_x}[x_tx_{t-1}^{\top}]$,
whereas RC-L-EM uses the regime-conditioned moments in
Eq.~\eqref{eq:rclem_conditional_cross_moment}.
The distinction therefore concerns the posterior moment
approximation, not analytic versus sampled evaluation alone.

\subsection{Switching Poisson cubature filtering and EM}
\label{app:spcf_baseline}

Song et al. developed regime-conditioned filtering and smoothing
for switching neural state-space models \citep{song2022modeling}.
The subsequent sPCF-EM method combines switching Poisson cubature
filtering, smoothing, and EM parameter learning
\citep{song2023unsupervised}. It uses cubature in the Poisson
measurement step, whereas RC-L-EM uses the local Laplace update
in Eq.~\eqref{eq:rclem_local_laplace}. Both use regime-conditioned
filtering rather than the global trajectory Laplace approximation
in Eq.~\eqref{eq:baseline_vlem_continuous}.

\section{Spatial pooling  obscures timescales}
\label{app:pooling}
In neural data analysis, some  methods pool spike counts across channels to reduce sparsity in short time bins of neural recordings. 
This operation can increase the number of events per bin in the accumulated one-dimensional time series, and thus it can facilitate autocorrelation analysis when individual channels are
sparse. However, it also changes how strongly each latent
dynamical mode contributes to the observed temporal structure. A pooled
signal may therefore reveal fewer timescales than are present in the
underlying population. Therefore, applying estimating methods on the pooled signal may underestimate the underlying timescale number and values. In contrast, using state-space model on the multivariate neural is not affected and can give more reliable timescale estimates. Here, we explain this effect through the geometry
of spatial pooling and illustrate its consequences for timescale recovery
and model-order selection.

\paragraph{Pooling reweights latent temporal modes.}
Consider independent stationary latent modes with variances $s_i$ and
autocovariances
$\operatorname{Cov}(x_{i,t+\ell},x_{i,t})
=s_i e^{-\ell\Delta t/\tau_i}$.
Let the observations be $y_t=Cx_t+\epsilon_t$, where $\epsilon_t$ is
independent of the latent process and uncorrelated across time.
For positive lags, the population autocovariance is
\begin{equation}
\Gamma_y(\ell)
=\operatorname{Cov}(y_{t+\ell},y_t)
=\sum_{i=1}^{H}s_i c_i c_i^\top e^{-\ell\Delta t/\tau_i},
\end{equation}
where $c_i$ is the $i$th column of $C$.
Pooling the observations into a scalar signal $u_t=w^\top y_t$ gives
\begin{equation}
\gamma_u(\ell)
=w^\top\Gamma_y(\ell)w
=\sum_{i=1}^{H}
\underbrace{s_i(w^\top c_i)^2}_{\text{pooled mode amplitude}}
e^{-\ell\Delta t/\tau_i}.
\label{eq:pool-scalar}
\end{equation}
Thus, a rank-one spatial projection preserves the decay constants of
contributing modes but changes their amplitudes according to their
alignment with $w$. A mode with $w^\top c_i=0$ disappears from the pooled
autocovariance, while a small nonzero projection makes it difficult to
resolve from finite data. Although a scalar ACF can contain multiple
timescales, their visibility depends on these relative contributions.

For channel summation, $w=\mathbf{1}$. Suppose the loadings satisfy
$c_{di}=\mu_i+\xi_{di}$, where the $\xi_{di}$ are independent across
channels, have zero mean, and have variance $\sigma_i^2$. Pooling $n$
channels then yields
\begin{equation}
\mathbb{E}\!\left[
\left(\sum_{d=1}^{n}c_{di}\right)^2
\right]
=n^2\mu_i^2+n\sigma_i^2.
\end{equation}
Modes with coherent, nonzero mean loadings therefore acquire a contribution
that grows quadratically with the number of pooled channels. Modes whose
positive and negative loadings largely cancel receive substantially less
weight. Consequently, increasing the total count level can reduce the
relative visibility of some latent timescales.

\paragraph{Empirical study.}
We design an expeirment to study 
the effect of spatial pooling in a stationary system with one regime ($K=1$). Threre are three independent  modes, which 
have unit stationary variance and timescales $(10,40,150)$ ms. We generate spike counts data that contains 120 trials of 1000 bins, with bin width $\Delta t=2$ ms
and 300 observed channels. A fixed loading matrix
$C\in\mathbb{R}^{300\times3}$ is drawn according to
$c_{di}=\mu_i+\xi_{di}$, with
$\mu=(0.30,0.05,1.00)$ and $\xi_{di}\sim\mathcal{N}(0,1)$.
Conditional spike counts are independent Poisson observations with means
\begin{equation}
\lambda_{d,t}
=\max\{b+\alpha C_d x_t,10^{-4}\},
\end{equation}
where $C_d$ denotes row $d$ of $C$.
We consider dense and sparse count settings with
$(b,\alpha)=(8,2)$ and $(0.25,0.045)$, respectively, using the same latent
trajectories and loading matrix. Their empirical means are 7.9311 and
0.2474 counts per bin per channel. Negative affine rates occur in
2.82\% and 0.82\% of entries, respectively, and are replaced by the
positive floor. The linear covariance calculation therefore provides an
approximate description of this rectified-rate simulation.

For each $n\in\{1,2,3,5,10,20,50,100,200,300\}$, we sample 30 subsets of
$n$ channels and sum their counts. All pooling levels use the same
recording within each count setting. We subtract the across-trial mean
at each time point and estimate the normalized ACF of the residual
pooled activity. One-, two-, and three-exponential mixtures are fitted
over positive lags from 2 to 298 ms by nonlinear least squares, with
nonnegative amplitudes, timescales greater than 1 ms, and 14 random
initializations per fit.

We assess timescale visibility using the fraction of channel draws for
which a three-exponential fit recovers each true timescale within 30\%
relative error. We also examine the selected exponential order using
\begin{equation}
\mathrm{BIC}(r)
=L\log\!\left(\frac{\mathrm{SSE}_r}{L}\right)+2r\log L,
\end{equation}
where $L=149$ is the number of fitted lags, $r$ is the number of
exponentials, and $\mathrm{SSE}_r$ is the residual sum of squares.
This criterion serves as a descriptive model-order comparison because
ACF errors at different lags are correlated. The reported fractions
summarize channel draws from the fixed recording; at full pooling,
all draws produce the same signal and fitting variation arises only
from optimization. To examine the consequences of fitting fewer
components, we additionally report the timescales and relative
amplitudes of the lowest-residual one- and two-exponential fits at
full pooling.

\begin{figure}[htp]
\centering
\includegraphics[width=\linewidth]{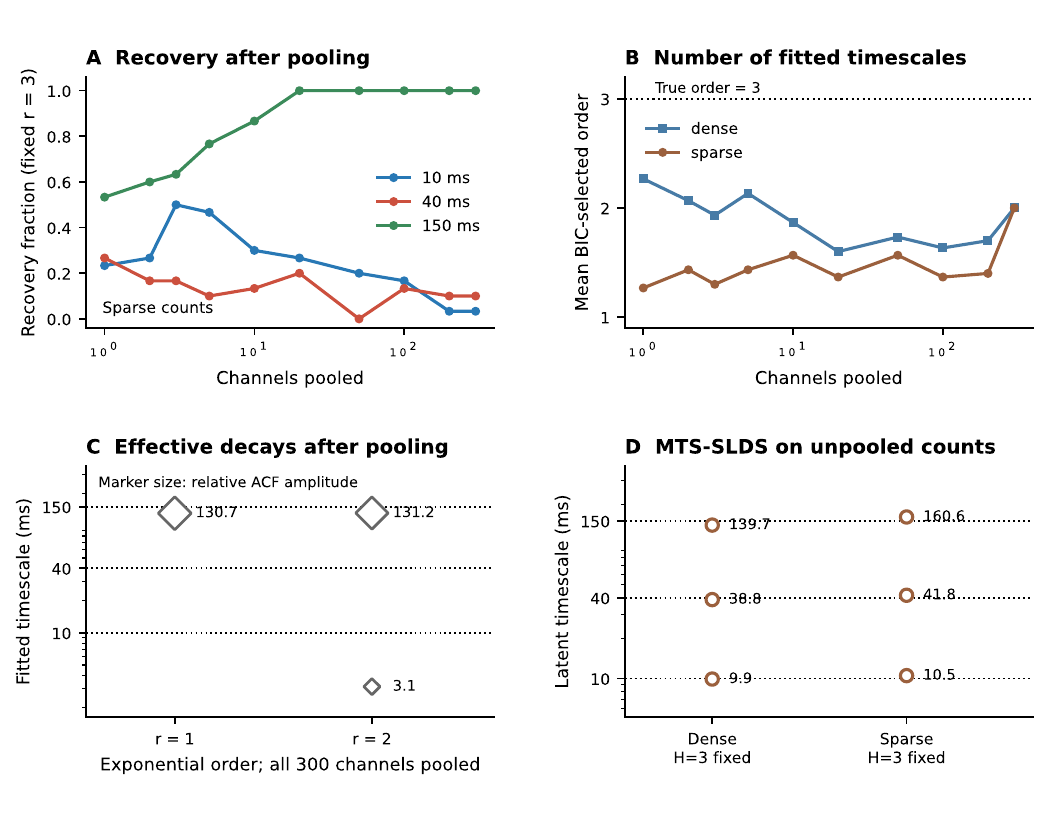}
\caption{\textbf{Pooling can obscure latent temporal modes in a single recording.}
\textbf{A} Per-mode recovery fractions from three-exponential ACF fits in
the  setting.
\textbf{B} Mean BIC-selected exponential order in both count settings.
A and B reuse the same generated recording at every pooling level, with 30
channel draws per level; full-pooling repetitions vary only optimization.
\textbf{C} Effective timescales of one- and two-exponential fits at full
pooling in the  setting. Marker size indicates relative
ACF amplitude; the lowest-residual fit at each order is displayed.
\textbf{D} 
Dotted lines indicate the true order or timescales.}
\label{fig:pooling}
\end{figure}

\paragraph{Consequences for timescale inference.}
Figure~\ref{fig:pooling} illustrates how pooling can reduce the visibility
of latent timescales despite increasing the count level. At full pooling,
BIC favors two exponentials in both count settings, although the
generating system contains three distinct timescales. This lower inferred
order reflects the temporal components resolved in the pooled statistic. Fitting fewer exponentials also changes the interpretation of the
estimated decay constants. A weak component may be omitted, or several
components may be represented by fewer effective timescales.  Pooling can therefore produce both an incomplete
timescale spectrum and estimates that summarize mixtures of latent
decays.

This limitation arises at the level of the observed representation and
also concerns aABCtau when inference is based on the ACF of a pooled
scalar signal \citep{zeraati2022flexible}. In contrast, 
multichannel state-space models such as MTS-SLDS retain channel-specific
loading patterns and cross-channel temporal dependencies. These provide
information about latent modes that may contribute only weakly to the
population sum and thus enable MTS-SLDS to provide more reliable timescale estimates.

\section{Additional synthetic experiments}

\subsection{Ablation tests}
\label{app:ablation}
MTS-SLDS combines multi-lag moment initialization with RC-Laplace-EM. In this section, we 
examine their contributions through several
ablation tests. For each test, we evaluate both timescale estimation and regime recovery.

\paragraph{Experimental setup.}
We generated Poisson observations from switching systems with $D=128$
observed dimensions, $H=10$ latent dimensions, $\Delta t=5$~ms, and
self-transition probability $p_{\mathrm{stay}}=0.9975$.
Each dataset contained 30 training and 20 held-out trials of 800 bins.
The $K=2$ and $K=3$ experiments used mean count levels of 0.2 and 0.5
counts per bin, respectively.
Initialization comparisons used ten seeds; comparisons involving
\texttt{ssm} fitting used five.
We ran RC-Laplace-EM for 300 iterations and \texttt{ssm}'s
\texttt{laplace\_em} for 100, reporting the final iterate without a
convergence-based stopping rule.
We computed timescale mean absolute percentage error (MAPE) over all
$KH$ modes after Hungarian matching in log-timescale space.
We aligned regime labels by posterior overlap, independently of
timescale matching, and evaluated held-out regime accuracy and minimum
per-regime recall.
Unless stated otherwise, numerical summaries are medians across seeds with interquartile ranges (IQRs).

\paragraph{Initialization under a common fitting procedure.}
For $K=2$, moment and \texttt{ssm} initialization reached similar final
timescale errors of approximately 9.5\% under the same RC-Laplace-EM
procedure (Fig.~\ref{fig:ablation_init_rc}A).
Moment initialization nevertheless reached 15\% MAPE in fewer iterations:
all ten runs crossed this threshold, compared with nine of ten
\texttt{ssm}-initialized runs.
Among runs that crossed it, the median first-crossing iteration was
10 (IQR 10-10) versus 130 (100-150); the remaining run was
right-censored at the 300-iteration budget.
Thus, the observed $K=2$ benefit concerns the optimization trajectory,
rather than a lower final error at this budget.

For $K=3$, initialization also affected final regime recovery.
Moment initialization with the ridge descriptor produced no
regime-recovery failures, whereas \texttt{ssm} initialization failed
in six of ten paired datasets, defining failure as minimum per-regime
recall below 0.10 (Fig.~\ref{fig:ablation_init_rc}B).
Final timescale MAPE was 9.8\% (IQR 9.5-10.6) versus
25.1\% (9.4-31.0).
The median paired difference, \texttt{ssm} initialization minus moment
initialization, was 14.9 percentage points (IQR 0.3-21.1;
Wilcoxon signed-rank test, $p=0.037$;
Fig.~\ref{fig:ablation_init_rc}C).
These results support the tested combination of moment initialization
and the ridge descriptor.
In this implementation, the dynamics update retains a regime's
parameters when its effective occupancy falls below one.
This safeguard can impede recovery of poorly occupied regimes, but
does not establish that regime loss is mathematically irreversible.

We next compared the two fitting configurations using moment-based
starts (Fig.~\ref{fig:ablation_init_rc}D).
RC-Laplace-EM reduced median MAPE from 19.1\% to
9.5\% (IQR 9.3-9.7), whereas Laplace-EM increased it
from 17.7\% to 56.2\% (54.1-100.1).
The latter fits had held-out regime accuracy of 0.500 and minimum
recall of zero on all five seeds.
The realized starting parameters were not identical.
The configurations also differed in iteration budget and emission
updates: the RC-Laplace-EM arm held the emission loading matrix $C$
fixed, whereas the \texttt{ssm} arm updated it.
Consequently, this comparison demonstrates different outcomes for
the tested fitting configurations; it does not isolate the effect
of RC inference or a single optimization update.

To isolate the effect of inference, we evaluated the same fitted
\texttt{ssm} parameters using its native posterior approximation and
the regime-conditioned filter, with identical initial-state
specifications (Fig.~\ref{fig:ablation_stickiness}A).
Across five seeds, held-out regime accuracy increased from
0.507 (IQR 0.488-0.649) to 0.770 (0.742-0.806).
MTS-SLDS achieved 0.949 (0.945-0.950).
Changing inference therefore recovered substantial regime information
without refitting the baseline parameters, although it did not close
the gap to MTS-SLDS.
The displayed trial was selected for the best MTS-SLDS recovery among
15 eligible trials from the seed with a saved posterior.
We use the all-seed summaries, rather than this illustrative trial,
to assess the comparison.

\begin{figure}
    \centering    \includegraphics[width=0.75\linewidth]{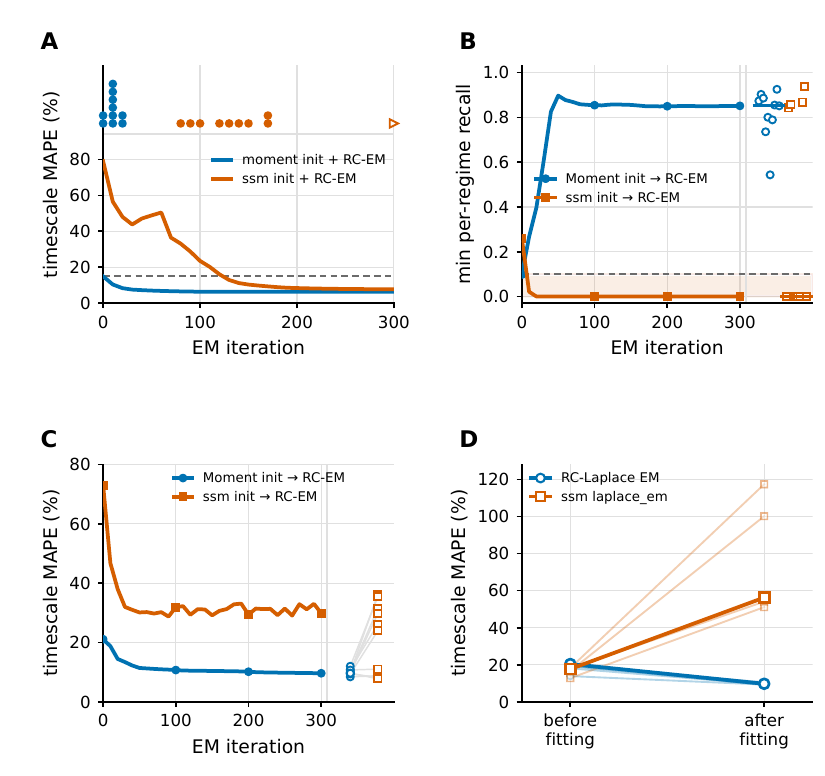}
    \caption{\textbf{Effects of initialization and fitting configuration.}
\textbf{A} Timescale MAPE for $K=2$ under RC-Laplace-EM with
moment or \texttt{ssm} initialization. The upper strip shows iterations
to reach 15\% MAPE across ten seeds (triangle, threshold not reached).
\textbf{B, C} Minimum per-regime recall and timescale MAPE for $K=3$
under the same initialization comparison.
Curves in \textbf{A-C} show example runs; right-hand points in
\textbf{B, C} show final outcomes across seeds.
\textbf{D} Timescale MAPE before and after RC-Laplace-EM or
Laplace-EM fitting from moment-based starts.
Thick lines highlight one seed; light lines show the others.}
\label{fig:ablation_init_rc}
    \label{fig:ablation_init_rc}
\end{figure}

\paragraph{Sensitivity to the transition prior.}
We varied the baseline's sticky transition prior across six
settings (Fig.~\ref{fig:ablation_stickiness}B).
A prior centered on the generating self-transition probability,
which uses oracle information, reduced median timescale MAPE from
31.2\% without a sticky prior to 15.2\%.
However, its native inference still gave regime accuracy of 0.507
and minimum per-regime recall of 0.019.
Matching the MTS-SLDS prior, centered at $p_{\mathrm{stay}}=0.99$,
gave 33.8\% MAPE, emphasizing that matching the methods' priors differs
from matching the generating process.
MTS-SLDS's median MAPE of 9.5\% remained below all six settings.
Thus, transition regularization improved parameter estimation,
but the absence of a sticky prior alone did not explain the
observed regime recovery gap.

\begin{figure}
    \centering
    \includegraphics[width=0.75\linewidth]{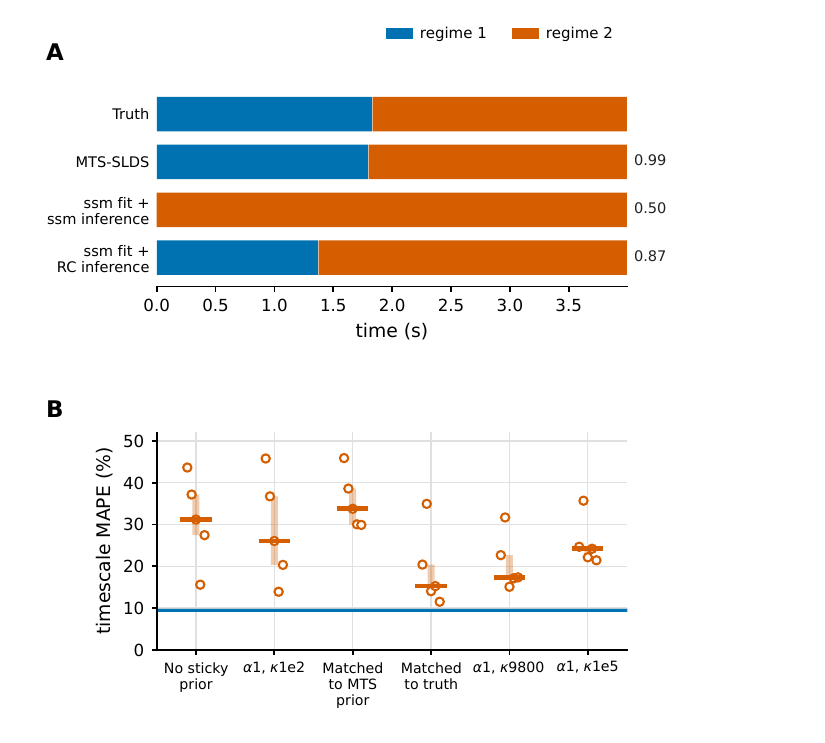}
    \caption{\textbf{Regime inference and sensitivity to transition priors.}
\textbf{A} True and inferred regimes for a trial in
 MTS-SLDS recovery among 15 eligible trials; numbers indicate
balanced accuracy. The two \texttt{ssm} rows use identical fitted
parameters with native or regime-conditioned inference.
\textbf{B} Timescale MAPE across 6 transition-prior
settings, including an oracle prior matched to the generating
transition probability.
Open circles show individual seeds ($n=5$), bars show medians,
and shading indicates interquartile ranges.
The blue line marks the MTS-SLDS median.}
\label{fig:ablation_stickiness}
    \label{fig:ablation_stickiness}
\end{figure}

\subsection{Recovery of oscillatory and decaying dynamics}
\label{app:oscillatory}
In the synthetic experiments of the main text, we report timescale recovery across many conditions. One important scenario of timescale anlaysis of real data is the case where there are not only decay time but also oscillations. Here we show that MTS-SLDS method can readily apply this case. 

\begin{figure}[t]
    \centering
    \includegraphics[width=0.8\linewidth]{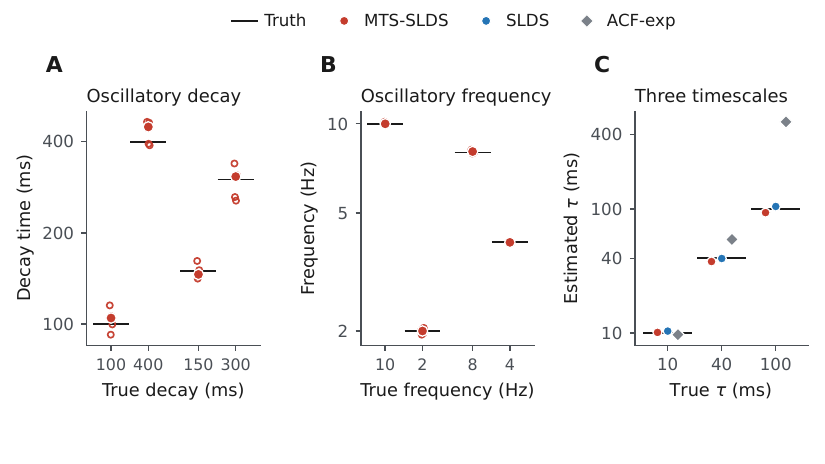}
    \caption{
    A,B Decay times and frequencies recovered by MTS-SLDS
    for oscillatory dynamics.
    C Recovery of three decay timescales by MTS-SLDS, SLDS,
    and ACF-exp.
    Black horizontal lines indicate ground truth, where red and blue
    markers denote MTS-SLDS and SLDS, respectively, and gray
    diamonds denote ACF-exp.}
    \label{fig:synthetic_additional_recovery}
\end{figure}

We evaluate MTS-SLDS on synthetic dynamics with damped
oscillations, testing recovery of both the decay timescale and
oscillation frequency.
Across the tested modes, with decay times of 100--400~ms and
frequencies of 2--10~Hz, MTS-SLDS returned estimates close to
ground truth (Fig.~\ref{fig:synthetic_additional_recovery}A,B).
Thus, the method recovered the decay envelope and oscillatory
frequency as distinct properties of the latent dynamics.

In a separate experiment with three decay timescales of
10, 40, and 100~ms, both MTS-SLDS and SLDS closely recovered
the true values (Fig.~\ref{fig:synthetic_additional_recovery}C).
ACF-exp overestimated the two slower components, particularly
the 100-ms timescale.
This comparison illustrates a limitation of direct exponential
ACF fitting in resolving multiple coexisting decay processes.

\section{V4 data analysis details}
\label{sec:v4_appendix}
We analyzed extracellular recordings from the 16 channels V4 data during
81 fixation trials, using a continuous 3-s epoch per trial and
2-ms bins (1500 bins per trial).
We treated these epochs as ongoing population activity rather than
an event-locked response and fitted all state-space models with
one dynamical regime ($K=1$).
A fixed trial-index split (seed 20260914) reserved 10 trials for
final testing; the remaining 71 trials formed the development set.

We fitted Poisson MTS-SLDS and the \texttt{ssm} reference
(SSM-SLDS) using five-fold cross-validation, three initialization
seeds, and latent dimensions $H\in\{2,4,6,8,10\}$.
We averaged predictive gains across seeds within each trial before
computing uncertainty, avoiding repeated counting of the same trial. 
The main comparison uses matched $H=8$, not independently selected
dimensions.

Both methods were evaluated through the same Laplace-approximate
filter using their fitted parameters.
We measured one-step predictive log-likelihood relative to a
constant-rate Poisson baseline estimated from the corresponding
training data.
The reserved test trials were not used for fitting or model selection,
and supplementary diagnostics used archived parameters without refitting.

For $H=8$, we grouped timescales across the 15 seed--fold fits
by single linkage with a 0.12 gap in log-timescale space.
We retained bands present in every fit and centered above one
sampling bin.
MTS-SLDS yielded band centers of 26.4, 49.5, and 120.2~ms;
SSM-SLDS yielded 23.5, 36.1, and 110.9~ms.
We reported one center per band rather than counting closely spaced
eigenvalues as separately resolved processes.
This criterion is conservative: every MTS-SLDS fit contained an
additional mode between 69.7 and 101.9~ms, but its variability prevented
any single sub-band from attaining full occupancy.

\begin{figure}[t]
    \centering
    \includegraphics[width=0.8\linewidth]
        {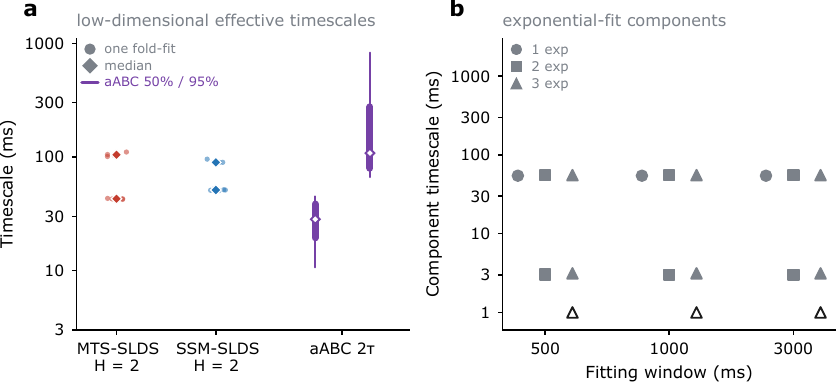}
    \caption{\textbf{Effective timescales and exponential-fit diagnostics in V4.}
    (a) Estimates from $H=2$ state-space models and two-timescale aABC.
    Points show fold fits, diamonds show medians, and aABC intervals.
    (b) Components from one-, two-, and three-exponential ACF fits
    across fitting windows.}
    \label{fig:v4_effective_timescales}
\end{figure}

We also compared pooled autocorrelations after summing channels,
subtracting the trial-average time course and grand mean, and
normalizing at zero lag.
Forward simulations from the state-space models underwent the same
processing.
aABC fitted this statistic over 0--100~ms; its agreement within that
interval therefore measures fit to its objective, not independent
predictive performance.
The state-space models had smaller ACF errors over 100--500~ms than
the two- and three-timescale aABC fits, with the latter using our
three-component extension.
Direct exponential fits provided a complementary diagnostic:
increasing their order introduced near-bin components and a
boundary-limited 1-ms component across the tested fitting windows
(Fig.~\ref{fig:v4_effective_timescales}b).

At matched $H=8$, MTS-SLDS improved predictive likelihood by
52.37~nats per trial, with positive differences on all 10 reserved
trials.
This corresponds to approximately 0.0022~nats per channel--bin
observation.
The advantage remained 44.14~nats per trial when compared with
SSM-SLDS at its development-optimal $H=4$.
A trial-level likelihood decomposition attributed the improvement
primarily to predicted firing rates, rather than covariance corrections.

\begin{figure}[t]
    \centering
    \includegraphics[width=0.8\linewidth]
        {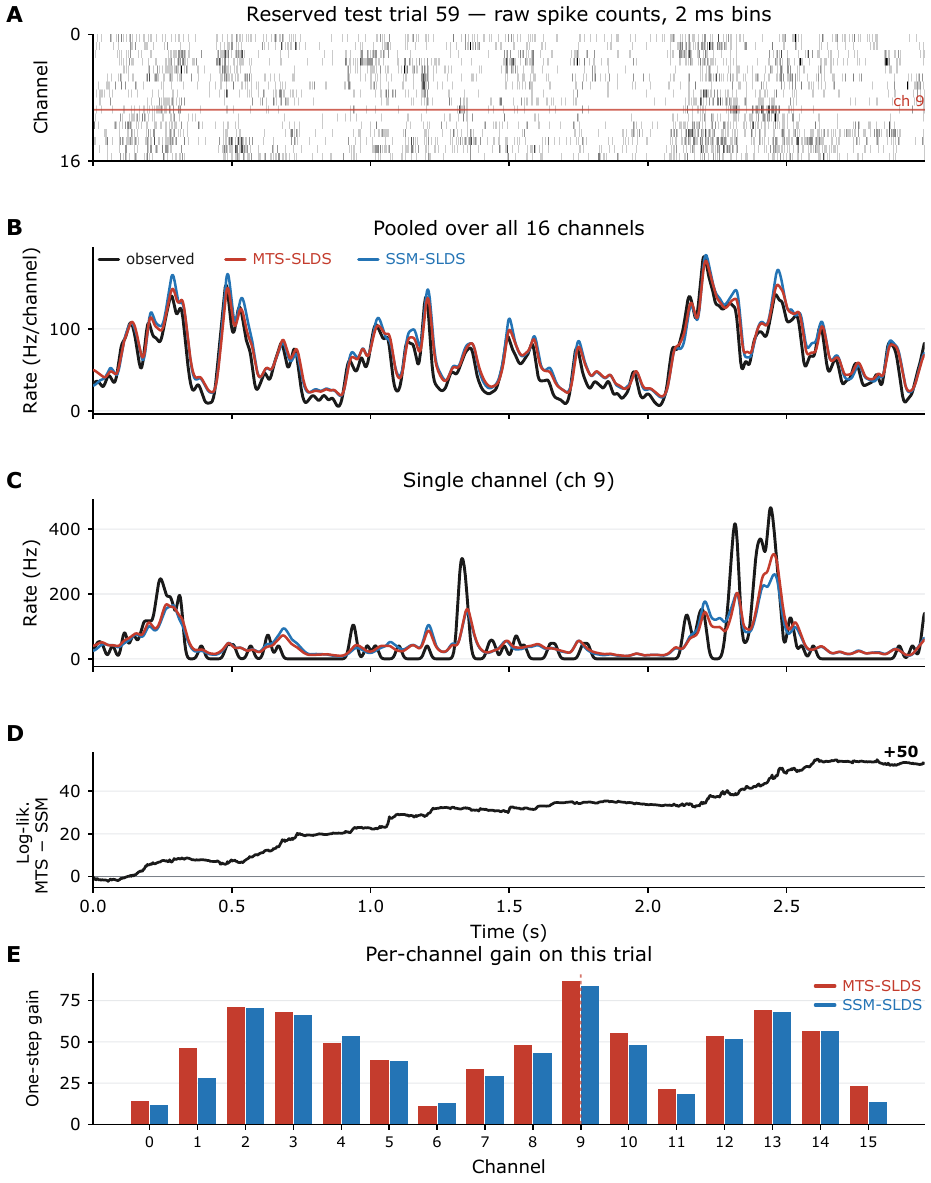}
    \caption{\textbf{Held-out V4 prediction at  $H=8$}.
    (A) Raw spike counts from the same trial 59.
    (B,C) Observed and predicted rates averaged across channels
    and for a selected channel example (channel 9).
    (D) Cumulative predictive log-likelihood difference
    between MTS-SLDS and SLDS.
    (E) Per-channel one-step predictive gains over the constant-rate Poisson baseline.}
    \label{fig:v4_channel_predictions}
\end{figure}

Figure~\ref{fig:v4_channel_predictions} shows population-averaged and
single-channel predictions, cumulative likelihood differences, and
per-channel gains for one reserved trial.
The single-channel trace is a selected best-channel example, not a
representative-channel summary.
Together, these analyses support multiple reproducible effective
timescale bands and improved held-out prediction, while leaving
the exact number and separation of decay processes dependent on
model dimension and spectral resolution.

In the main text, we select latent dimension through the prediction performance. Now we want to study how the timescales look like with specifically lower latent dimension, as according to \citep{zeraati2023intrinsic} it is reported that two timescales beat three timescales for this dataset. In MTS-SLDS, we can force the model to learn two timescales by setting $H=2$.
Setting $H=2$ restricts the model to at most two distinct spectral
decay times, but it does not prescribe their values.
These low-dimensional fits yielded effective components in the
tens-of-milliseconds and approximately 100-ms ranges, and they reside in estimates of  the
two-timescale aABC fit, while the later method showed broader posterior uncertainty
(Fig.~\ref{fig:v4_effective_timescales}a).
Meanwhile, ACF-exp fits tends to give much faster timescales when using two and three exponentials (Fig.~\ref{fig:v4_effective_timescales}b).

\section{Area 2: fitting and analysis}
\label{sec:area2_appendix}

\paragraph{Data and model fitting.}
We analyzed 88 trials, where 22 trials are in active reaches and 66 in passive arm
perturbations.
We binned spike counts at 20~ms over a window from $-500$ to
$+500$~ms relative to movement onset and fitted each condition separately.
We evaluated Poisson MTS-SLDS and SLDS 
using five repeated 90/10 training/test splits, $K=2$ regimes, and
$H\in\{4,6,8,10,12\}$ latent dimensions.
MTS-SLDS used multi-lag moment initialization followed by
regime-conditioned Laplace-EM; SLDS used PCA-ARHMM initialization and variational Laplace-EM.
Neither model used hand kinematics as inputs.
Increasing $H$ from 4 to 6 improved held-out predictive gain over a
constant-rate Poisson baseline for both methods in both conditions;
subsequent increases gave inconsistent gains.
We therefore used $H=6$ for the displayed comparisons
(Fig.~\ref{fig:area2_main}B).

\begin{figure}
    \centering
    \includegraphics[width=0.8\linewidth]{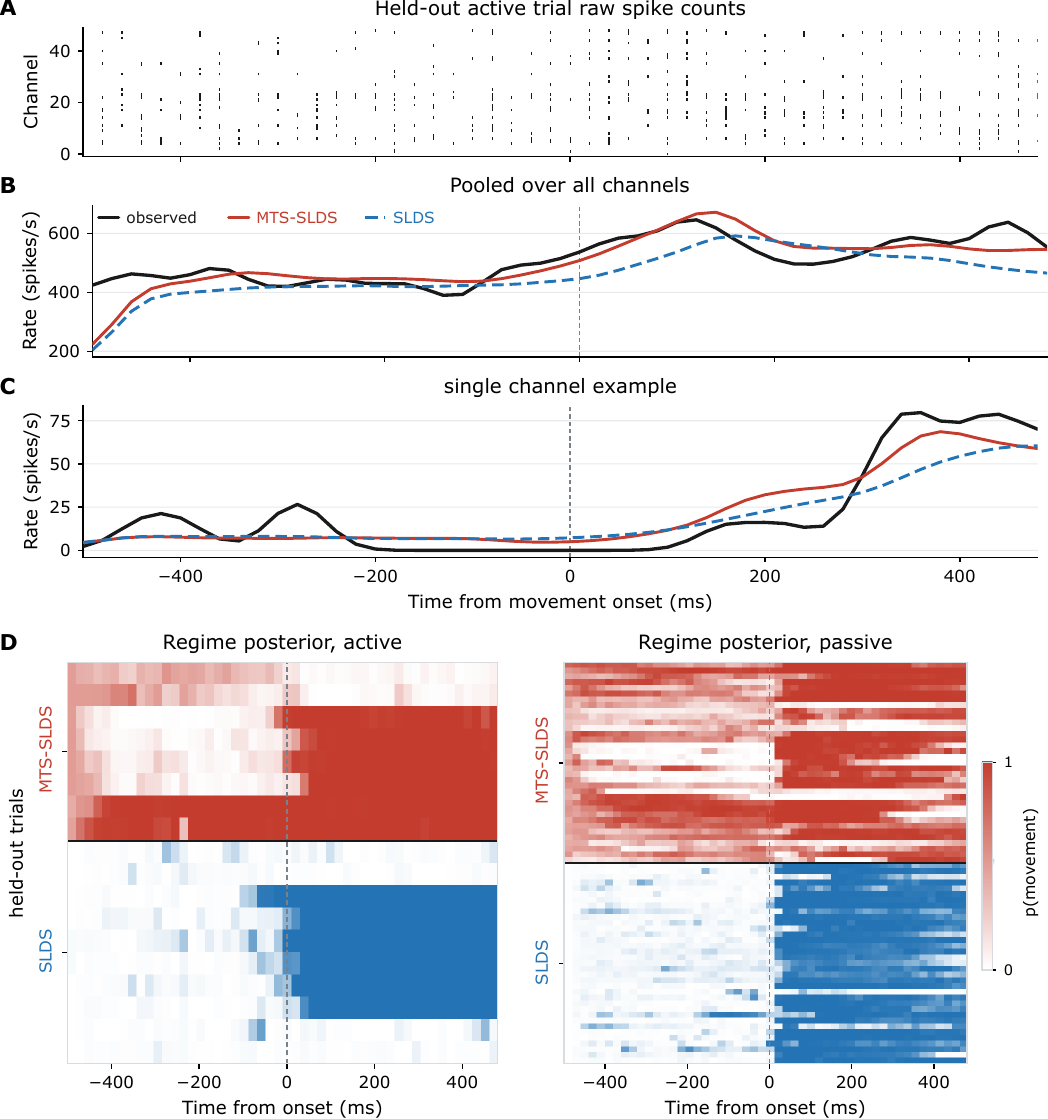}
    \caption{\textbf{Held-out trial activity prediction and regime inference in Area 2.}
\textbf{A} Spike counts from an example held-out active trial.
\textbf{B,C} Observed smoothed firing rates (black) and estimates from
MTS-SLDS (red) and SLDS (blue dashed), pooled across channels B and for a single channel C.
\textbf{D} Movement-regime posterior probabilities across held-out
trials during active reaching (left) and passive perturbations
(right), where each row represents one trial and vertical dashed lines mark movement onset.}
\label{fig:area2_diagnostics}
\end{figure}

\paragraph{Regime inference and held-out predictions.}
Both methods identified changes in regime probability near movement
onset (Fig.~\ref{fig:area2_main}C).
The trial-level posteriors also reveal substantial heterogeneity:
MTS-SLDS assigned appreciable movement-regime probability before onset
on a subset of trials, particularly during passive perturbation
(Fig.~\ref{fig:area2_diagnostics}D).
For an illustrative held-out active trial, we show raw spikes,
population-pooled rates, and single-channel one-step predictions
(Fig.~\ref{fig:area2_diagnostics}A-C).
Each prediction used observations only through the preceding bin;
Gaussian smoothing was applied to the observed rate for display.
The single channel was selected by its MTS-SLDS predictive score and
then shown for both methods, so this panel illustrates a favorable
example rather than typical single-channel performance.

\paragraph{Spectral summaries and temporal control.}
We extracted timescales from stable transition eigenvalues using
$\tau=-\Delta t/\log|\lambda|$.
We retained out-of-band modes in the spectrum plot but restricted
numerical summaries to 20-300~ms, a practical analysis range given
the bin width and short regime durations, rather than a formally
established identifiability interval
(Fig.~\ref{fig:area2_main}D).
Panel D summarizes in-band modes by their geometric mean; panels E
and G use the slowest retained mode in each regime and split.
The paired comparisons include three active and four passive splits.
The apparent pre-movement to movement shifts at $H=6$ did not remain
consistent across band choices and latent dimensions
(Fig.~\ref{fig:area2_main}E).
For the temporal control, we permuted population count vectors across
time within each trial and refitted the models, preserving count
distributions and within-bin cross-channel structure.
The median ratio of real to shuffled geometric-mean in-band timescales
was 1.9 for MTS-SLDS and 1.4 for SLDS, based on five and four paired
split summaries, respectively (Fig.~\ref{fig:area2_main}F).

\paragraph{Behavioral comparison.}
We summarized the falling phase of hand speed by the
exponential-equivalent decay time $\tau_{\mathrm{beh}}=t_{1/2}/\ln 2$.
Median values were 168~ms for 19 active trials and 86~ms for 66 passive
trials, a 1.95-fold difference.
Median slowest in-band movement-regime timescales were 188 and 99~ms
for MTS-SLDS, compared with 117 and 108~ms for SLDS
(Fig.~\ref{fig:area2_main}G).
The MTS-SLDS estimates thus matched the behavioral magnitudes and
ordering more closely, but different emission settings across
conditions confound attribution of this contrast to latent dynamics.
Moreover, an exploratory test relating trial-level regime occupancy
to movement duration found no evidence of association
($\rho=0.12$-$0.17$, $n=35$, $p>0.34$).
This test concerns regime occupancy, not trial-specific dynamical
timescales. The repeated training/test splits also overlap and do not
constitute independent biological replicates.
We therefore interpret the comparison as agreement in magnitude,
without establishing neural-behavioral coupling or separating
intrinsic from input-driven dynamics.

\paragraph{Separate sPCF-EM method analysis.}
We also examined a separate timescale analysis using 
sPCF-EM method \citep{song2023unsupervised}, with $H=6$,
$K=2$, and three folds per condition
(Fig.~\ref{fig:area2_spcf}).
The timescales patterns found here are different from MTS-SLDS and SLDS, and is less aligned with the behavior timescales as well.

\begin{figure}
    \centering
    \includegraphics[width=0.8\linewidth]{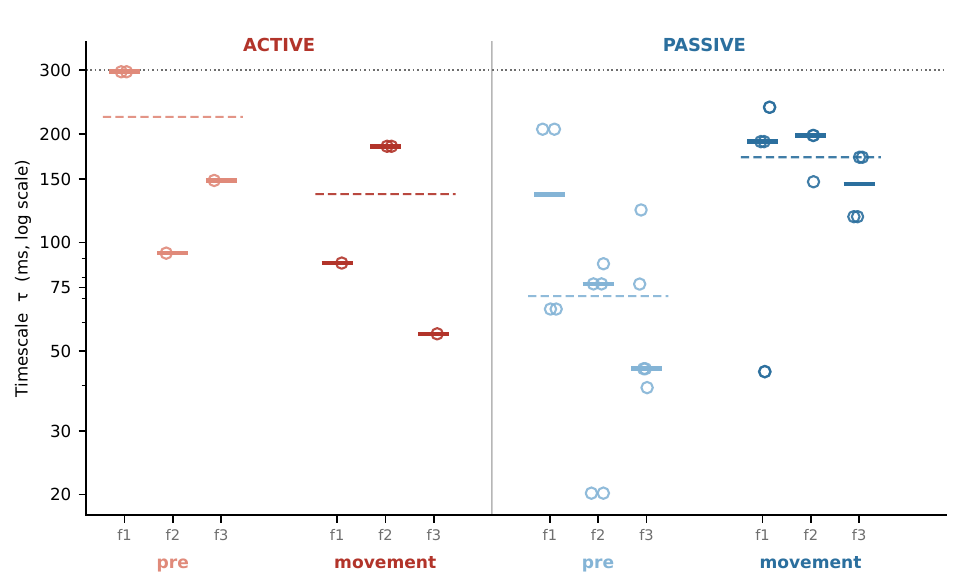}
    \caption{Timescales estimated by sPCF-EM are shown for pre-movement and movement regimes during active  (red) and passive (blue) conditions.}
\label{fig:area2_spcf}
    \label{fig:area2_spcf}
\end{figure}

\end{document}